\documentclass{ieeeaccess}
\usepackage{cite}
\usepackage{amsmath,amssymb,amsfonts}
\usepackage{algorithmic}
\usepackage{graphicx}
\usepackage{textcomp}
\usepackage{float}
\usepackage{lineno}
\usepackage{caption}
\usepackage{subcaption}
\usepackage{tabularx}
\usepackage{array}
\usepackage{booktabs}
\usepackage{amsmath}

\def\BibTeX{{\rm B\kern-.05em{\sc i\kern-.025em b}\kern-.08em
    T\kern-.1667em\lower.7ex\hbox{E}\kern-.125emX}}
\begin{document}
\history{Date of publication xxxx 00, 0000, date of current version xxxx 00, 0000.}
\doi{10.1109/ACCESS.2017.DOI}

\title{Efficient Scopeformer: Towards Scalable and Rich Feature Extraction for Intracranial Hemorrhage Detection}
\author{\uppercase{Yassine Barhoumi}\authorrefmark{1}, 
\uppercase{Nidhal C. Bouaynaya}\authorrefmark{2}, and \uppercase{Ghulam Rasool}.\authorrefmark{3}}
\address[1]{Electrical and computer science department,Rowan University, NJ, United States (e-mail: barhou29@students.rowan.edu)}
\address[2]{Electrical and computer science department,Rowan University, NJ, United States (e-mail: bouaynaya@rowan.edu)}
\address[3]{Machine Learning Department, Moffitt Cancer Center, Tampa, FL, United States (Ghulam.Rasool@moffitt.org)}


\markboth
{Author \headeretal: Preparation of Papers for IEEE TRANSACTIONS and JOURNALS}
{Author \headeretal: Preparation of Papers for IEEE TRANSACTIONS and JOURNALS}

\corresp{Corresponding author: Yassine Barhoumi (e-mail: barhou29@students.rowan.edu).}

\begin{abstract}
The quality and richness of feature maps extracted by convolution neural networks  (CNNs) and vision Transformers (ViTs) directly relate to the robust model performance. In medical computer vision, these information-rich features are crucial for detecting rare cases within large datasets. This work presents the "Scopeformer," a novel multi-CNN-ViT model for intracranial hemorrhage classification in computed tomography (CT) images. The Scopeformer architecture is scalable and modular, which allows utilizing various CNN architectures as the backbone with diversified output features and pre-training strategies. We propose effective feature projection methods to reduce redundancies among CNN-generated features and to control the input size of ViTs. Extensive experiments with various Scopeformer models show that the model performance is proportional to the number of convolutional blocks employed in the feature extractor. Using multiple strategies, including diversifying the pre-training paradigms for CNNs, different pre-training datasets, and style transfer techniques, we demonstrate an overall improvement in the model performance at various computational budgets. Later, we propose smaller compute-efficient Scopeformer versions with three different types of input and output ViT configurations. Efficient Scopeformers use four different pre-trained CNN architectures as feature extractors to increase feature richness. Our best Efficient Scopeformer model achieved an accuracy of 96.94\% and a weighted logarithmic loss of 0.083 with an eight times reduction in the number of trainable parameters compared to the base Scopeformer. Another version of the Efficient Scopeformer model further reduced the parameter space by almost 17 times with negligible performance reduction. In summary, our work showed that the hybrid architectures consisting of CNNs and ViTs might provide the desired feature richness for developing accurate medical computer vision models.\end{abstract}



\maketitle

\section{Introduction}
\label{sec:introduction}
Stroke is a leading cause of death globally \cite{Elliot-2010-intracerbral-hemorrhage}. The seriousness of this condition makes the early detection of brain hemorrhages vital for patient prognosis. Early detection and classification of intracranial hemorrhaging using head computed tomography (CT) scans can reduce the risk of serious medical complications and extensive brain damage, particularly within the first 24 hours \cite{Elliot-2010-intracerbral-hemorrhage,Wardlaw-2001-Cochrane-thrombolysis}. Early detection and classification often require qualified physicians to evaluate and detect any indications of bleeding inside the cranium or the existence of a lesion within the brain tissues. Machine learning algorithms trained to autonomously identify and classify brain hemorrhages can decrease the detection time, allowing quicker and more effective treatment. Emerging computer vision techniques offer faster and more robust models that can triage patients and help expert physicians and radiologist efficiently use their time \cite{Gong-2007-brain-Trauma-classification,Chilamkurthy-2018-Retrospective-study-head-CT,Ajay-2019-3D-CT-hemorrhage}. 


In recent years, Convolutional Neural Networks (CNNs) and Vision Transformers (ViTs) have received significant attention in the field of computer vision for their ability to process and analyze large amounts of visual data. CNNs utilize convolution operations and pooling layers to extract discriminative and meaningful features from an image. These features are then used to classify images or recognize objects. On the other hand, ViTs incorporate self-attention mechanisms, borrowed from transformers in NLP, to dynamically weigh the importance of each part of an image when making predictions. This approach has proven effective in image classification and object recognition tasks, demonstrating the potential of ViTs to solve more complex computer vision problems.

Motivated by the performance of CNNs and ViTs, we propose a hybrid architecture consisting of multiple CNNs and a multi-encoder ViT model. The model, which we called Scopeformer, accommodates ($n$) numbers of the CNNs dedicated to the extraction features, and several stacked ViT encoders dedicated to differentially extracting weights from the global feature map. These weights represent inter-feature correlations learned by the model as relevant for the hemorrhage classification problem. The results show that the classification accuracy is proportional to the number of CNNs used to extract the features in the Scopeformer training, leading to higher computational requirements. In this work, we present the Scopeformer model and investigate selective feature engineering methods to generate richer content for classification. We also address the large trainable parameter space issue presented by increasing the number of CNNs for feature extraction. We employ dimensionality reduction techniques on the convolution feature space to control the attention complexity within the ViT encoders. Our experiments resulted in building a scalable and efficient hybrid multi-convolution-based ViT model (n-CNN-ViT) to solve the hemorrhage detection problem. Our contributions can be summarized as:

\begin{itemize}
    \item Proposed a hybrid architecture called Scopeformer, combining multiple CNNs and a multi-encoder ViT model for intracranial hemorrhage detection in CT images.
    \item Results showed that the classification accuracy was proportional to the number of CNNs used for feature extraction, leading to higher computational requirements.
    \item Addressed the issue of large trainable parameter space by employing dimensionality reduction techniques on the convolution feature space to control the attention complexity within the ViT encoders.
    \item Developed a scalable and efficient hybrid multi-convolution-based ViT model for the hemorrhage detection problem, resulting in improved classification accuracy.
\end{itemize}


\section{Background and Literature Review}
\subsection{RSNA Intracranial Hemorrhage Detection}
In 2019, the Radiological Society of North America (RSNA) provided a large number of brain CT scans of healthy participants and patients with internal cerebral hemorrhage of various types. RSNA held a machine learning challenge to foster the development of autonomous algorithms for multi-class hemorrhage classification \cite{Flanders-2020-RSNA-dataset}. The computerized multi-label classifiers were designed to determine whether there was cerebral bleeding in each 2-dimensional (2D) slice of the input CT image and to give a probability vector with six components relative to classification targets.

\subsection{Convolution Neural Networks (CNNs)}
Until recently, in computer vision applications, convolutional neural networks (CNNs) have been the de facto models for extracting high-resolution features for downstream tasks, e.g., classification \cite{Burduja-2020-ICH-classification,DelRocini-2020-abnormalities-CT-scans,trustworthy, extended, brain, premium}. The official top-ranking solutions for the RSNA challenge, posted on the Kaggle online community platform, employed multi-stage classification models incorporating convolution-based feature extraction stage \cite{RSNA-top-solutions}. Different stacking and arrangements of convolutional layers yield different features. These features are subject to implementation variations of various configurations, such as the architecture structure, the parameters governing the visual information flow, and the depth of the model \cite{Aubry_2015_ICCV}. Several off-the-shelf architectures were proposed based on broadening the perceptual field, improving the feature extraction efficacy, and reducing the trainable parameter space for faster and efficient computation \cite{EfficientNet,ResNet,DenseNet,Xception, VGG}. Increasing the depth of the architecture better approximates the target and provides improved feature representations due to the higher non-linearity and the improved receptive field. Enhancements were based on including and optimizing the design of the convolution layers, activation functions, loss functions, regularization methods, and optimization processes \cite{Advances-in-CNNs}.



\subsection{Vision Transformers (ViTs)}
Vision Transformers (ViTs) are increasingly being employed in a wide range of computer vision identification applications \cite{Distillation-Transformer, Swin-Transformer} and have proven successful in a multitude of vision tasks such as the ImageNet classification challenge \cite{ImageNet-2012}. The basic working component of ViTs is the Transformer block \cite{transformers}, originally introduced by Vaswani et al. \cite{Attention-is-all-you-need} in the realm of the Natural language process (NLP). The successful implementation of the Transformer model \cite{Attention-is-all-you-need} applied to images, known as vision Transformer or ViT, was a milestone in the computer vision field \cite{Dosovitskiy-2021-ViT}. Various successful implementations of the ViT architecture in the medical field were proposed that outperformed the standard convolution-based models by a large margin \cite{Dai-2021-multi-modal-medical-classification}. ViT model \cite{Dosovitskiy-2021-ViT} divides a natural image into equal 3-channel square patches. These patches are flattened and represent uni-dimensional tokens. Each patch represents local semantic information of the raw image, and the model learns to extract patterns from their correlations. Finer patches result in the extraction of higher local correlations and improved semantics due to the large quadratic complexity of the model. However, this complexity results in expensive computations and large data requirements. It was shown that such models only outperform standard CNNs in high data regimes in either pre-training or training \cite{Dosovitskiy-2021-ViT}. 

\subsection{Convolution Neural Networks against Vision Transformers}

Convolutional Neural Networks (CNNs) and Vision Transformers (ViT) are two models for image recognition in computer vision, with different approaches to feature extraction. CNNs use convolutional layers to learn local patterns, resulting in a pyramid-like structure\cite{Yamashita-2018-Convolution-overview}, while ViT uses self-attention mechanisms to process global patterns, resulting in a columnar structure \cite{Zhang-2021-PVT}. CNNs handle image scale through pooling layers and are faster and more memory-efficient, while ViT uses a linear projection followed by self-attention to learn a fixed-size representation of the image, making it more effective in learning global patterns and flexible in terms of input size. 

While CNNs \cite{ResNet, Identity-mappings} integrate local information derived from input images by combining multiple convolutional operations, ViTs \cite{Dosovitskiy-2021-ViT} learn patterns from spatial information and non-local dependencies exploiting the encoder block's multi-head self-attention (MHSA) function \cite{Attention-is-all-you-need}. These patterns allow ViT models to gain increasingly rich global context without manually constructing layer-wise local characteristics. Applying attention to all pixels in an image increases the impact of global feature correlations, which allows the model to derive more relevant hidden patterns. As shown in \cite{Distillation-Transformer}, stacking multiple ViT encoders tends to increase the model performance, and with the appropriate training methods, a model constructed of 12 blocks of ViT encoders outperformed a ResNet model consisting of more than 30 bottleneck convolutional blocks on the ImageNet classification task \cite{Distillation-Transformer}. However, it is shown that increasing the depth of ViTs via stacking Transformer blocks does not necessarily increase model performance \cite{Zhou-2021-DeepViT}. In fact, ViT model performance plateaus and starts declining beyond certain numbers of stacked encoders \cite{Zhou-2021-DeepViT}. Zhou et al. \cite{Zhou-2021-DeepViT} identified an attention collapse issue and proposed a new mechanism, termed \emph{Re-attention}, that accounts for correlations among the attention multi-heads. The proposed model entitled DeepViT presented delayed plateauing behavior that enables more block stacking to achieve higher performance. 

In our previous publication \cite{Barhoumi-2021-Scopeformer}, we introduced the Scopeformer, a hybrid architecture combining the strengths of convolutional neural networks (CNNs) and a vision transformer. This architecture was designed to extract high-level features from various inputs. The Scopeformer takes advantage of the ability of CNNs to capture local patterns in data and the ability of the vision transformer to capture global dependencies and relationships between different parts of the input. Combining these two architectures results in an effective and efficient model capable of performing complex feature extraction tasks. The introduction of the Scopeformer represents a significant step forward in the field and opens up new avenues for future research and development.



\section{Materials and Methods}
This paper presents the Scopeformer, a hybrid multi-CNN vision transformer model, and its improved version, the Efficient Scopeformer. The model processes pseudo-RGB CT scan images through CNNs and a vision transformer to extract high-level features for classification. Our focus is to improve performance and efficiency through modifications to the architecture. The results of these modifications are analyzed and presented in this paper.
\subsection{Scopeformer}
\begin{figure}[H]
\centering
\includegraphics[width=9 cm]{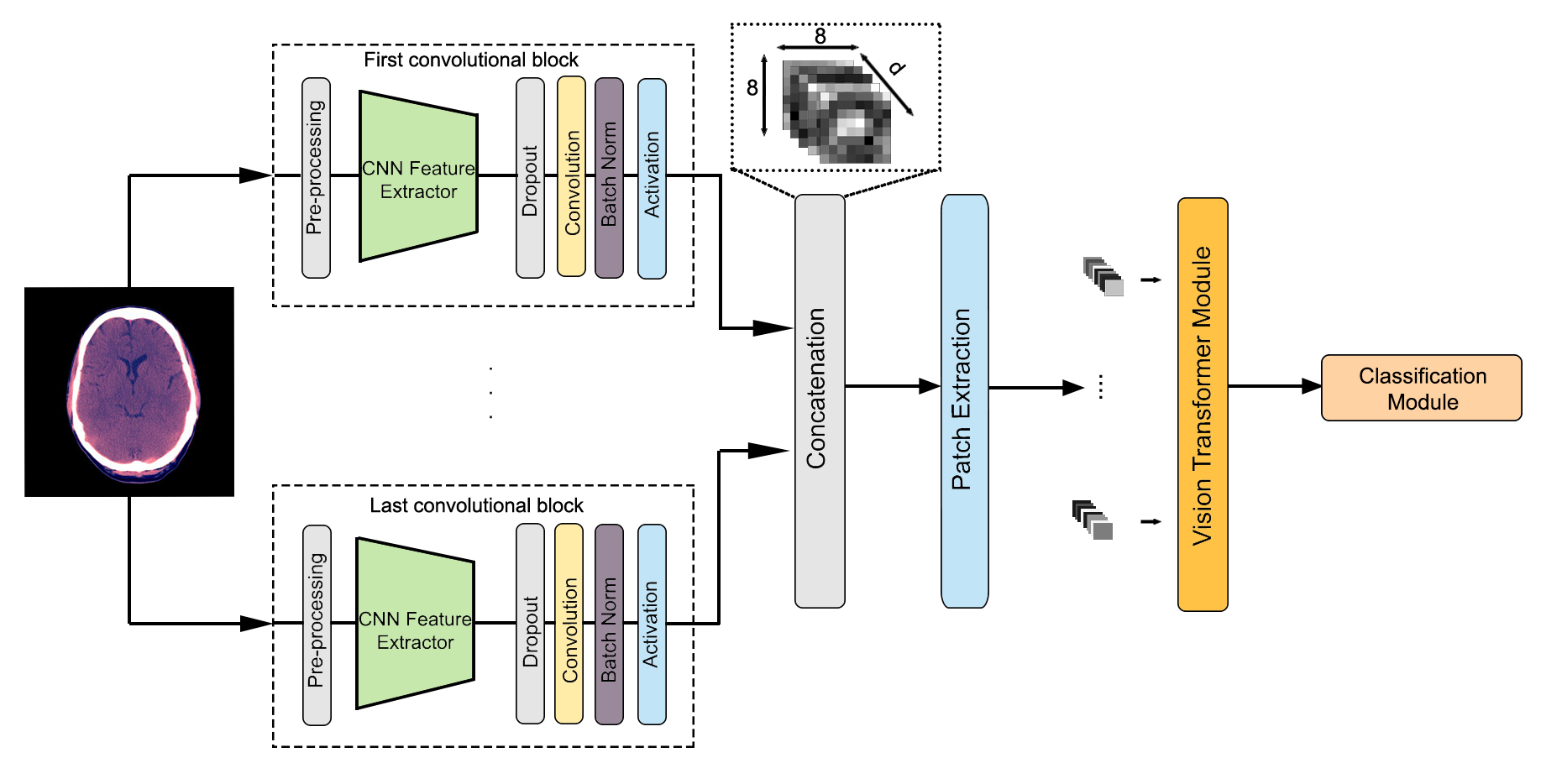}
\caption{A schematic layout of the Scopeformer architecture is presented. The proposed model is composed of four main modules: (1) Scopeformer Backbone, (2) patch extraction, (3) vision Transformer (ViT) encoder, and (4) classification head. A single input image is fed to several CNN models to extract various features and construct feature maps. These feature maps are processed by the patch extraction module and vectorized. The vectors form the input to the Transformer encoder, and the model output is taken from the classification module.\label{fig2}}
\end{figure}

We present our hybrid n-CNN-ViT model in Figure \ref{fig2}. The model is composed of $n$ number of CNN models stacked to build the feature-extractor backbone.

We refer to the n-CNN-ViT model as  ``Scopeformer'', derived from the ``Transformer'' (-former) and the word "Scope-" for the \emph{selective feature extraction backbone} generated from the convolution blocks with deep receptive fields. The Scopeformer model brings significant advancement in ViTs and CNNs. The main difference between a Scopeformer model and ViT resides in employing high-level features with more semantic information as input to the Transformer encoder, as opposed to the originally proposed ViT model, which inputs raw natural images in the form of small patches. The Scopeformer model takes global convolutional feature maps in the form of smaller but deeper patch sizes. The ViT patch extraction method divides a natural image into patches along the height and width, then flattens every patch and joins all channels into a single 1-D token. Similarly, we pixel-wise divide the feature map along the height and width of the features into $p$ $\times$ $p$ patches, where $p$ is the feature-patch size (with $p$ = 1 for all the experiments in this work).

The input to the model consists of a tensor with a dimension of $H$ $\times$ $W$ $\times$ $C$, where $H$ represents the height, $W$ represents the width, and $C$ is the number of concatenated channels derived from the RSNA DICOM files. The model executes a concurrent forward pass of the input images through different CNN architectures and stores the output features $f$. These features are concatenated along the channel axis. The resultant global feature map has a dimension of $h$ $\times$ $w$ $\times$ $c$, where $h$ represents the features height, $w$ represents the features width, and $c$ is the total number of features with $c$ = $n$ $\times$ $f$. 

The first Scopeformer architecture uses Xception CNNs \cite {Xception} and several ViT layers. The Xception model comprises several Inception modules composed of depth-wise and point-wise convolutions. In our Scopeformer model, we stack ($n$) differently pre-trained Xception models \cite {Xception} in the feature extraction backbone and freeze updates on their weights during training. We use the last inception layers embedded within the Xception models as feature generators. The ImageNet pre-trained Xception CNNs, present high-level features to the ViT block. To this end, we consider that the primary role of the ViT block is to extract \emph{correlations} from depth-wise patches. The global feature map can be generated using one or more Xception blocks stacked in the same Scopeformer as depicted in Fig. \ref{fig2}. Our initial experiments consider stacking raw features from CNN blocks without any further processing.


\begin{figure}[htbp]
\begin{minipage}[b]{.3\linewidth}
  \centering
  \centerline{\includegraphics[width=2.5cm]{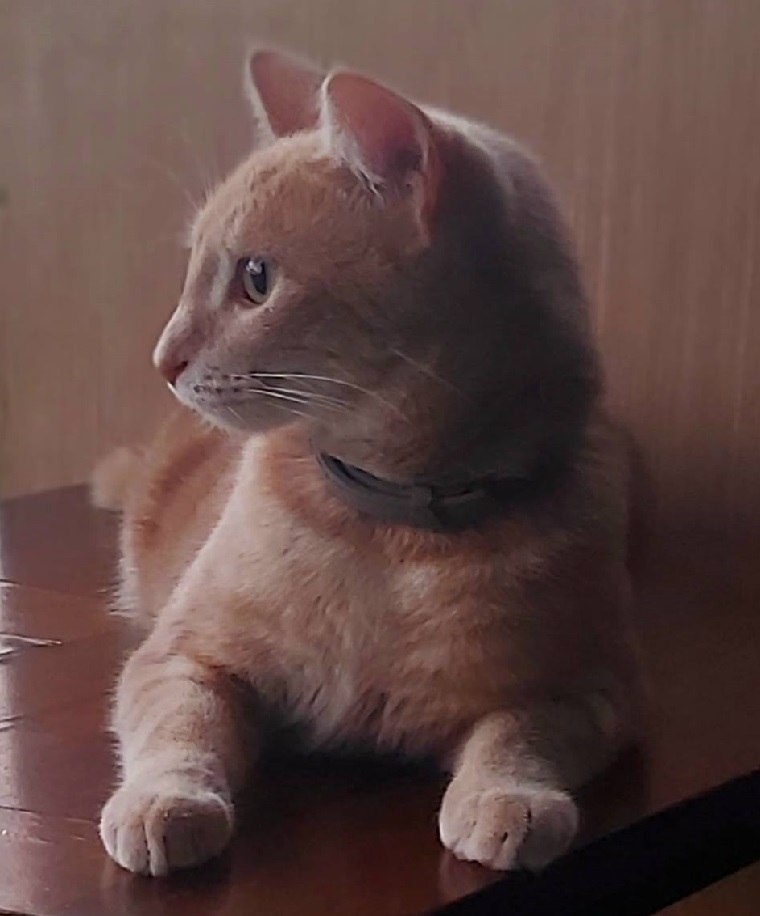}}
  \centerline{(a) Content image}
\end{minipage}\hfill
\begin{minipage}[b]{.3\linewidth}
  \centering
  \centerline{\includegraphics[width=2.5cm]{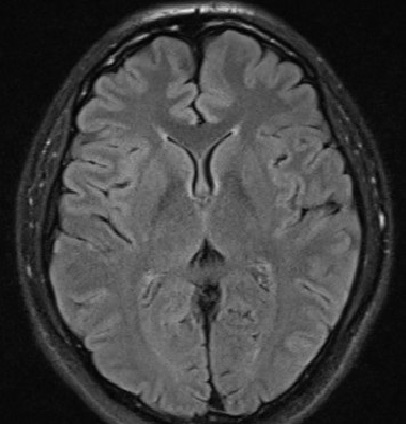}}
  \centerline{(b) Style image}
\end{minipage}\hfill
\begin{minipage}[b]{.3\linewidth}
  \centering
  \centerline{\includegraphics[width=2.5cm]{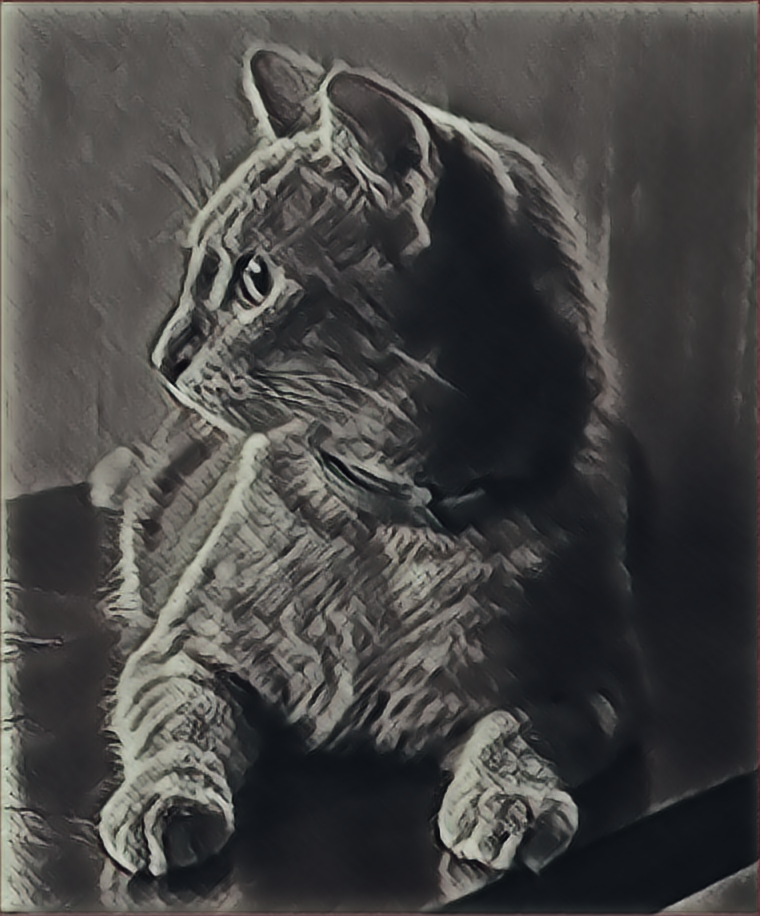}}
  \centerline{(c) Output image}
\end{minipage}\hfill

\caption{Style transfer method applied on ImageNet dataset. (a) Content image, (b) Style image, and (c) Output image}
\label{Style-transfer-Cat}
\end{figure}

In our formulation, we tend to diversify the pre-training methods of every Xception CNN. This allows for generating different features specific to each architecture. In the first phase of model training, we load the ImageNet pre-trained weights in all CNNs using Keras API \cite{Keras}. In the second phase of training, Xception CNNs are trained to perform different classification tasks, including the RSNA hemorrhage dataset to perform classification. We used hard data augmentation on one of the CNNs and soft data augmentation on the others. We applied style transfer \cite{Style-Transfer} on the ImageNet dataset to induce a grayscale brain-like image shape bias as depicted in Fig. \ref{Style-transfer-Cat}. The output dataset was used to pre-train the third CNN. In our experiments, we tested several combinations of the pre-trained CNNs within the Scopeformer architecture. 

\subsection{Efficient Scopeformer}
After extensively testing the Scopeformer model, we formulated and included several innovations in the feature extractor CNNs and the ViT blocks. We define four modules as presented in Fig. \ref{fig2}. The first module is the Scopeformer Backbone and represents the stack of multiple CNNs contributing to the global feature map. The second module is designed for patch extraction (from the CNN features) to generate ViT tokens. The third module consists of the ViT pipeline. Finally, the fourth module represents the classification head. We discuss these modules in the following sections.

\subsubsection{Module 1: Scopeformer Backbone}
Efficient Scopeformer uses a variety of CNNs to build the feature extraction block. The backbone CNNs include ImageNet-pre-trained ResNet 152 V2 \cite{ResNet}, EfficientNet B5 \cite{EfficientNet}, DenseNet 201 \cite{DenseNet}, and Xception \cite{Xception}. The features generated by each CNN are concatenated along the channel axis to form a \emph{global feature map}. However, constructing such a feature map requires that the individual feature maps generated by each CNN have identical height and width. We propose augmenting each CNN with a single trainable $ 1 \times 1 $ convolutional layer that projects the features to an appropriate space.
The input to the Efficient Scopeformer consists of a tensor with a dimension of $H$ $\times$ W $\times$ $3$, where $H$ represents the height, $W$ represents the width, and $3$ is the number of channels. The image is concurrently fed to four CNNs to generate high-level feature maps. The channel dimension of all four feature maps will be reduced using $1 \times 1$ convolution layer to $8 \times 8 \times \frac{d}{4}$, where $d$ is the size of the \emph{global feature map}.

\subsubsection{Module 2: Patch extraction}
The input dimension of the second module depends on the size of the \emph{global feature map} set by the first module. In our experiments, the resultant  \emph{global feature map} is a 3D tensor with a shape of $8 \times 8 \times d$. The patch extraction module splits the features across the height and width channel-wise and extracts $ N = \frac{8\times 8}{p^2}$ $d$-dimensional vectors. We set the patch size to $1 \times 1$ and get $N = 64$ tokens representing one local pixel position of features across all the $d$ features. The dimension d is controlled by the projection method used in the previous module and represents a bottleneck of the architecture. Every patch contains semantic information on the local pixel position across all the generated features from the four CNNs. The resultant sequence of flattened patches $X_p \in \mathbb{R}^{64 \times d} $ is then used as the input set for the ViT block.

\subsubsection{Module 3: Scopeformer ViT}
\begin{figure}[htbp]
\centering
\includegraphics[width=9cm]{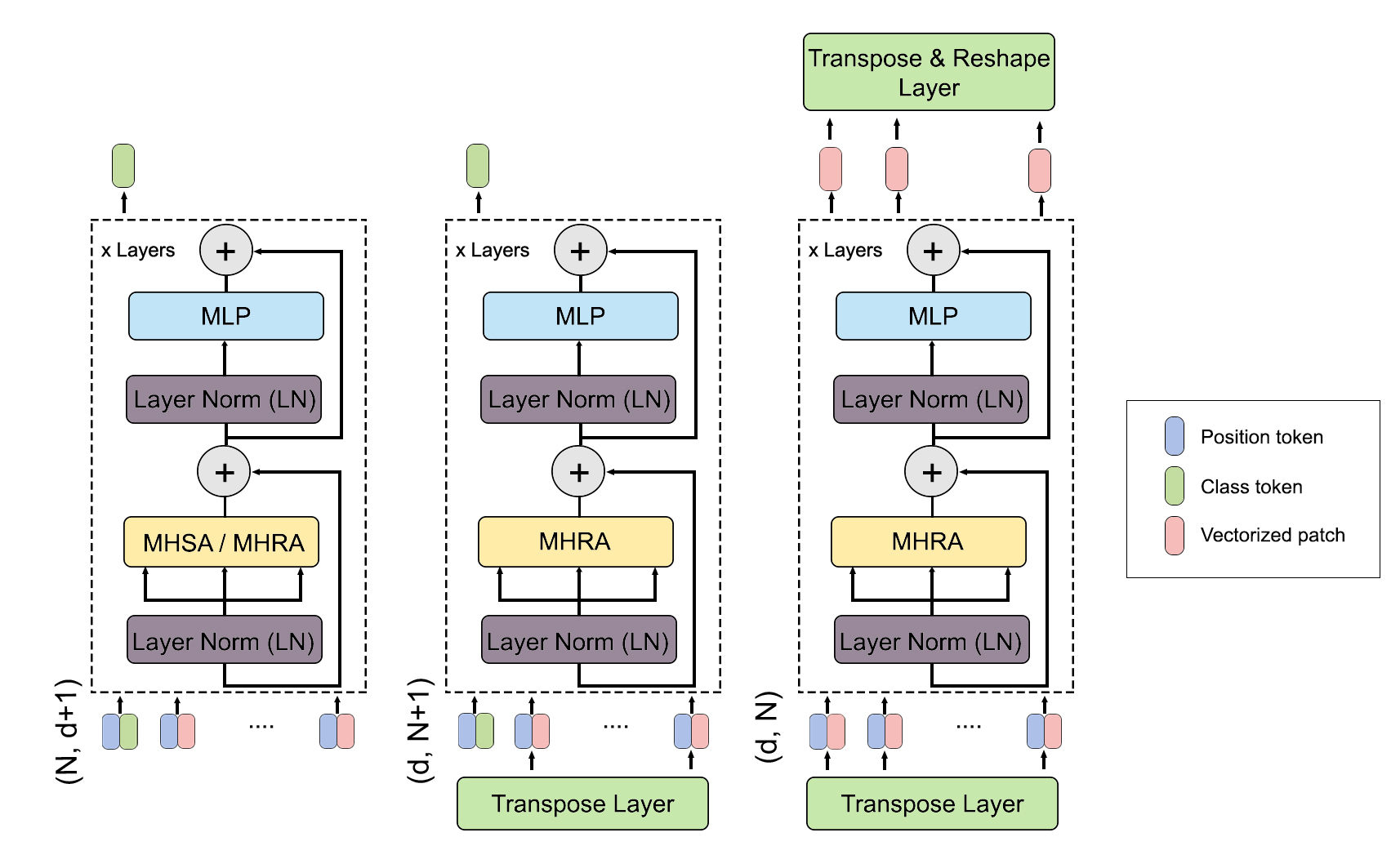}
\caption{ViT Scopeformer configurations. (Left) Baseline Scopeformer Configuration: The first configuration is a ViT block with an input of vectorized patches extracted from the CNNs features. \textbf{(Center) Deep Scopeformer TR Configuration:} The second configuration introduces a transpose layer to transform the channel-wise patches into feature-wise patches. \textbf{(Right) Efficient Scopeformer Configuration:} The third configuration dismisses the token class and uses all the feature tokens as input. The output of the third block will be transposed to retrieve back the dimension of the CNN features, which we feed to the classification module.}\label{fig:fig1}
\end{figure}

We evaluated three different ViT configurations for the proposed architecture as depicted in Figure \ref{fig:fig1}. These configurations include (1) Deep Scopeformer, (2) deep Scopeformer TR (Transpose), and (3) Efficient Scopeformer.

\unskip

\textbf{Baseline Scopeformer Configuration:} In this configuration, we feed a set of vectors generated by the patch extraction layer to ViT encoders. We used trainable position encoding vectors coupled with vectorized patches and a trainable class (CLS) token. The dimension of the input to the ViT encoder block is  $Y \in \mathbb{R}^{N \times d+1}$. We used two self-attention variants. The first one is referred to as multi-head self-attention (MHSA) \cite{Dosovitskiy-2021-ViT} and the second variant as the multi-head re-attention (MHRA) \cite{Zhou-2021-DeepViT}. The key difference resides in the introduction of a trainable transformation matrix. These variants are given by:
\begin{linenomath}
\begin{align}
    \text{MHSA}(Q, K, V) &= \text{Softmax}\left(\frac{QK^T}{\sqrt{d_k}}\right)V, \label{eq:MHSA} \\
    \text{MHRA}(Q, K, V) &= \text{Norm}\left(M^T\left(\text{Softmax}\left(\frac{QK^T}{\sqrt{d_k}}\right)\right)\right)V, \label{eq:MHRA}
\end{align}
\end{linenomath}

where $M \in \mathbb{R}^{h \times h} $ is a learnable transformation matrix, and $h$ is the number of self-attention heads.

\textbf{Deep Scopeformer TR Configuration:}
The second Scopeformer ViT configuration applies a transpose operation to the set of vectors produced by the patch extraction layer. The output of the transpose layer is summed up with the position-encoded vectors and concatenated with the CLS token. The dimension of the resultant set of vectors is $Y_T \in \mathbb{R}^{d \times N+1}$. We used only the MHRA self-attention variant (Eq. \ref{eq:MHRA}) in our experiments.

\textbf{Efficient Scopeformer Configuration:}
The third Scopeformer module discards the CLS notion used in previous configurations. In these settings, we use all the features generated by ViT encoders for classification. As such, the dimension of the input and output of ViT encoders remain identical and equal to $Y_T \in \mathbb{R}^{d \times N}$. We use a Transpose and Reshape layer at the ViT output to get the appropriate dimension for the feature map. We use the MHRA self-attention variant to compute self-attention.

\subsubsection{Module 4: Classification module}
The classification module in baseline and the deep Scopeformer TR configurations receives a single CLS token. The output of this token is turned into a prediction using a multi-layer perceptron (MLP) with a sigmoid activation function and a single hidden layer. In the efficient Scopeformer configuration, the classification module receives a set of reshaped features $x_t \in \mathbb{R}^{8 \times 8 \times d} $. The classification module applies a 2D average pooling layer, followed by a flattened layer. Finally, the class inference is made via a dense layer with a sigmoid activation function.


\subsection{Dataset}

The RSNA dataset was collected by Adam et al. \cite{Flanders-2020-RSNA-dataset} from multiple scanner types used in different institutions worldwide. The dataset is considered the current largest dataset publicly available, aimed to capture complex real-world details of the hemorrhage subtypes. The dataset was publicly released in the 2019 Intracranial Hemorrhage (ICH) detection challenge hosted by the Kaggle platform. The dataset contains 870,301 annotated 16-bit grayscale computer tomography (CT) scans saved in the DICOM format, annotated with five types of hemorrhage. Trained physicians categorized each CT slice with one or more types of a brain hemorrhage. Five different forms of hemorrhages are to be identified in this competition, with an additional class representing the presence of any hemorrhage type in the provided slice. These classes were labeled as Epidural hemorrhage (EDH), Intraparenchymal hemorrhage (IPH), Intraventricular hemorrhage (IVH), subarachnoid hemorrhage (SAH), and Subdural hemorrhage (SDH). 
\subsection{Pre-processing}

\begin{figure}[htbp]
\begin{minipage}[b]{.5\linewidth}
  \centering
  \centerline{\includegraphics[width=2.5cm]{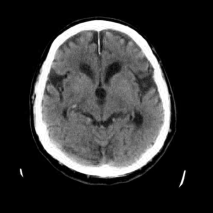}}
  \centerline{(a) Brain Tissue}
\end{minipage}\hfill
\begin{minipage}[b]{.5\linewidth}
  \centering
  \centerline{\includegraphics[width=2.5cm]{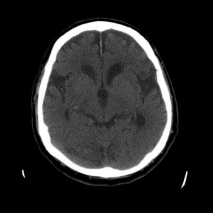}}
  \centerline{(b) Subdural}
\end{minipage}\hfill
\begin{minipage}[b]{.5\linewidth}
  \centering
  \centerline{\includegraphics[width=2.5cm]{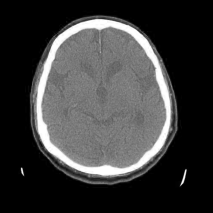}}
  \centerline{(c) Soft Tissue}
\end{minipage}\hfill
\begin{minipage}[b]{.5\linewidth}
  \centering
  \centerline{\includegraphics[width=2.5cm]{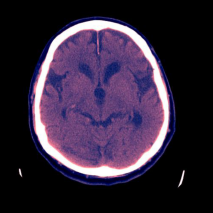}}
  \centerline{(d) Stacked Pseudo-Image}
\end{minipage}\hfill
\caption{Hounsfield unit CT slice conversion and the corresponding stacked 3-channel image.}
\label{HU-conversion}
\end{figure}

Individual images consist of pixels that have a range of 0 to $2^{16}$ with a resolution of $256^{2}$, referred to as Hounsfield Units (HU) \cite{Burduja-2020-ICH-classification}. HU represents the density of the scanned matter. Attenuation HU values are indicative of the content of the scan \cite{Broder-2011-Diagnostic-imaging}. For instance, bones have an attenuation value ranging between 250 and 1000, and fat and muscle have attenuation values (AV) ranging between 50 and 100. The RSNA CT scan DICOM files provide tags in the metadata about Hounsfield ranges used during registration of the CT scan. We use these tags to ensure standardization of the ranges across the dataset before applying HU windowing \cite{Recommendations-CT}. We use three windows of HU as channels in the input of the Scopeformer model, as depicted in Fig. \ref{HU-conversion}. Our settings for HU windows were: brain AV$\in[40,80]$ HU, subdural window AV $\in[80,200]$ HU, and soft tissue window AV $\in[40,380]$ HU \cite{Burduja-2020-ICH-classification}. 

\begin{table}[htbp]
\caption{Various design configurations of Scopeformer - hyperparameters and learnable parameters.}
\label{table-various-Scopeformer-designs}
\resizebox{0.47\textwidth}{!}{
\begin{tabularx}{0.6\textwidth}{X X X X X X X}
\toprule

\textbf{Model} & \textbf{CNN Blocks} & \textbf{Layers} & \textbf{Feature Size} & \textbf{MLP} & \textbf{Heads} & \textbf{Parameters} \\ 
\midrule
{Scopeformer (S)}  & 4 & 8               & 516            & 3072         & 12    & 34 M    \\ 
{Scopeformer (B)}  & 4 & 8               & 512            & 4096         & 16         & 42 M   \\ 
{Scopeformer (M)} & 4 & 8               & 512            & 5120         & 16      & 43 M  \\ 
{Scopeformer (L)/4} & 4 & 4               & 1024           & 4096         & 16      & 51 M   \\ 
{Scopeformer (L)/8} & 4 & 8               & 1024           & 4096         & 16      & 102 M  \\ 
{Scopeformer (L)/16} & 4 & 16               & 1024           & 4096         & 16     & 203 M \\   
{Deep Scopeformer (L)/8} & 4 & 8               & 1024           & 4096         & 16      & 102 M  \\ 
{Deep Scopeformer TR (L)/8} & 3 & 8               & 384           & 4096         & 16      & \textbf{6 M}  \\ 
{Efficient Scopeformer} & 3 & 8               & 384           & 4096         & 16      & \textbf{6 M}  \\

{Scopeformer \cite{Barhoumi-2021-Scopeformer}} & 3 & 12              & 3072            & 3072         & 8      & 755 M  \\ 
{Scaled Scopeformer \cite{Barhoumi-2021-Scopeformer}}  & 4 & 8               & 4096       & 4096  & 16  & 870 M \\    
\bottomrule
\end{tabularx}}
\end{table}


\subsection{Design Configurations and Experiments}
 Details about the various Scopeformer hyperparameter configurations and architectures are presented in Table \ref{table-various-Scopeformer-designs}. We present the different proposed Scopeformer variations and details about the number of convolution models used in the feature extraction backbone, the number of ViT layers, the global feature map size, the MLP dimension and the number of heads in each ViT block, and the total number of trainable parameters. We compare our Efficient Scopeformer implementation to our initial model implementation and propose lower trainable parameter space given the configuration hyperparameters. Our experiments comprise four main parts.

In the first set of experiments, we evaluate the size effect of various variants of Scopeformer on the classification accuracy, where four variants are evaluated; small (S), base (B), medium (M), and large (L). We keep the number of ViT layers fixed (equals 8) and increase the complexity of the model by configuring the MLP size residing in the ViT blocks for S, B, and M variants and increasing the feature size for the L variant. The number of trainable parameters drastically increases from the smallest (S) to the largest (L) variants.

In the second set of experiments, we investigate the effect of the number of ViT encoder blocks on the model performance. Based on preliminary results conducted in the first set of experiments, we conduct our ablation study on the large Scopeformer variant (L) with a feature size of 1024 and an MLP dimension of 4096. We consider three experiments where we gradually stack in an end-to-end fashion 4, 8, and 16 ViT encoders, forming three models named Scopeformer (L)/4, Scopeformer (L)/8, and Scopeformer (L)/16, respectively. Given the largest model parameters reside within the ViT architecture, the total number of trainable parameters is linearly scaled to the number of ViT blocks we use. 

The third set of experiments examines the transition from the originally proposed ViT model \cite{Dosovitskiy-2021-ViT} to a different version called DeepViT \cite{Zhou-2021-DeepViT}. We test this configuration on the highest performing model from the previous two sets of experiments; Scopeformer (L)/8 with a global feature map size of 1024, 8 layers of ViT encoders, and an MLP dimension of 4096. The model version, entitled Deep Scopeformer (L)/8, has a slightly higher number of trainable parameters.

The final experiment introduces three different ViT configurations to our Scopeformer architecture as depicted in Figure \ref{fig4}. We add these configurations to the highest-performing model from the previous three parts of the study; Deep Scopeformer (L)/8 with a global feature map size of 1024, 8 layers of ViT encoders, and an MLP dimension of 4096. We introduce and compare a set of three Scopeformer configurations, as presented in \emph{section 2.2.3}; Baseline Scopeformer configuration, Deep Scopeformer-TR configuration, and Efficient Scopeformer configuration.

\subsection{Pre-Training Efficient Scopeformer}
In all the experiments, we initially pre-trained the Scopeformer model using the ImageNet-1k dataset \cite{Russakovsky-2015-ImageNet}. Later, we train all models using the RSNA dataset \cite{Flanders-2020-RSNA-dataset}. In the first module (convolutional backbone), we freeze $ \approx 70\% $ of the layer weights in each CNN and keep top $ \approx 30\% $ trainable along with the newly introduced $ 1 \times 1 $ convolution layer. In our last experiment using the Efficient Scopeformer model, we pre-trained the backbone neural network on the RSNA dataset for hemorrhage classification for 150 epochs on top of the defaulted pre-training on ImageNet-1k. In this experiment, denoted as Efficient Scopeformer (p), we freeze weights of the feature extraction block during training. 

\subsection{The Loss Function}
Following guidelines from the RSNA Intracranial Haemorrhage Challenge (ICH), we adopted a weighted version of the \emph{multi-label logarithmic loss} function for our model training. The weighting was introduced to amplify the importance of classifying the first class representing all types of hemorrhages, with a coefficient of 2, at the expense of the rest of the classes, which have coefficients of 1. The evaluation of the loss value with respect to a single instance represents the weighted average over all the binary losses computed on each class individually. The ICH represents a multi-label classification problem, i.e., the input image can be classified into multiple classes, using binary labeling for each class to indicate its presence or absence. In our formulation, we applied multi-label hot encoding on the dataset to assign a binary value on each class for every CT slice. The \emph{multi-label logarithmic loss} function is defined as follows:

\begin{linenomath}
\begin{equation}
L_{\text{multi-BCE}} \left(y,\tilde{y}\right) = -  \sum_{n=1}^{6} \alpha_{\text{n}} (y_n \log \tilde{y_n}  + \left(1 - y_n\right) \log\left(1 - \tilde{y_n}\right)),
\label{eq3}
\end{equation} 
\end{linenomath}

where $\alpha_{\text{n}}$ represents the coefficient of the target classes, $y_n$ represents the ground-truth of each class n, and $\tilde{y_n}$ is the corresponding predicted probabilities.

\subsection{Evaluation Metrics}
The official model evaluation metric in the RSNA IHC was the \emph{weighted accuracy}. We evaluate the overall performance of the models based on three metrics, (1) the classification accuracy on the RSNA dataset, (2) the visual evaluation of the global feature richness of the embedding layer generated by the convolution backbone, and (3) the ratio of the model size function to the total number of trainable parameters. 

\section{Results and Discussion}

\subsection{The Effect of Backbone Neural Network Model Size and Pretraining Techniques}
We gradually stack $n$ various pre-trained Xception models in the feature extraction backbone. We freeze all these architectures in the backbone to prevent updates on their weights during training. We pre-trained the CNN models on diversified pre-training schemes, including ImageNet-1k natural image dataset (I) and the generated style transfer-base dataset (S). Table \ref{Scopeformer_results} compares different models and the corresponding performances on the hemorrhage classification task. While the n-CNN-ViT models were trained on the convolution features generated by the convolution backbone, the ViT model was trained on the raw dataset. The input dimension of the ViT block represents the full-resolution image or the set of features before splitting into patches. Results show that extracting features using convolution models to train the ViT model is a better alternative to the raw dataset. The Scopeformer model exploits the pre-training for generating high-level features useful for the ViT architecture. The use of CNNs leverages the need for high data regimes since the ViT model is used to fit these high-level features and extract semantic correlations instead of learning the spatial features in training. Furthermore, results show that the classification accuracy is proportional to the number of CNN models used in the Scopeformer training, i.e., as we stack feature extraction architectures in the backbone of the model, we get higher performances on hemorrhage classification. We further boost these performances by selectively varying the pre-training methods for each CNN architecture. We hypothesize that increasing the feature map size in the ViT input allows for increased semantic correlation extractions by the ViT block. Furthermore, diversifying the inductive biases derived from differently pre-trained CNN architectures may lead to a different set of feature maps, which contributes to a richer feature map and leads to observed improved performances. 

\begin{table}[htbp]
\caption{Classification performance of ViT-based models on the RSNA validation dataset. (I) represents pre-training on ImageNet. (S) represents pre-training on ImageNet with style transfer images.}
\begin{tabularx}{0.45\textwidth}{X X X X}
\toprule
\textbf{Model} & \textbf{ ViT input dimension}  & \textbf{Validation accuracy} & \textbf{Loss} \\ \midrule
\textbf{ViT}               & 256×256×3      & 94.33\%          & 0.1822       \\
\textbf{1-CNN-ViT (S)}     & 7×7×1024       & 96.95\%          & 0.08272      \\ 
\textbf{2-CNN-ViT (I-I)}   & 7×7×2048       & 97.22\%          & 0.07984      \\ 
\textbf{2-CNN-ViT (S-S)}   & 7×7×2048       & 97.26\%          & 0.07934      \\ 
\textbf{2-CNN-ViT (I-S)}   & 7×7×2048       & 97.46\%          & 0.07754      \\ 
\textbf{3-CNN-ViT (I-I-S)} & 7×7×3072       & 98.04\%          & 0.07050      \\  
\bottomrule
\end{tabularx}
\label{Scopeformer_results}
\end{table}
\unskip

\subsection{The Effect of Scopeformer Size}
Tables \ref{tab2} and \ref{tabindivtargets} show the results of experiments performed with different variants of the Scopeformer model. Table \ref{tabindivtargets} depicts different results obtained on individual classes of the S, B, and M models. We propose four sizes of the Scopeformer model, S, B, M, and L, with a reduced number of trainable parameters compared to our initial implementation of the Scopeformer model involving several Xception-based CNNs. The key component to the parameter reductions is linked to the trainable \emph{$ 1 \times 1 $ convolutional layer} placed after each convolution architecture in the feature extraction backbone before concatenation. In this experiment, we gradually increase the model complexity of S, B, and M variants by varying the MLP dimension and the number of self-attention heads within the ViT module as depicted in table \ref{table-various-Scopeformer-designs}. 

\begin{table}[htbp]
\caption{Performance of the different Scopeformer variants.}
\begin{tabularx}{0.45\textwidth}{X X X X X}
\toprule
\textbf{Model} & \textbf{ Accuracy }  & \textbf{Loss} & \textbf{Recall} & \textbf{Trainable Parameters} \\ \midrule
\textbf{Small (S)} & 93.00\%          & 0.1703          & 84.95\%  & 34M \\
\textbf{Base (B)} & 93.92\%       & 0.1461         & 89.29\% & 42M        \\ 
\textbf{Medium (M)} & 93.88\%        &  0.2285       & 88.44\%  & 43M  \\ 
\textbf{Large (L)/4}  & 93.12\%       & 0.1378        & 87.81\%  & 51M\\ 
\textbf{Large (L)/8} & \textbf{94.69}\%         & \textbf{0.1197}       & \textbf{89.33\% }   &102M \\ 
\textbf{Large (L)/16} & 92.57\%           & 0.1395        & 87.34\%  & 203M\\ 
\bottomrule
\end{tabularx}
\label{tab2}
\end{table}
\unskip

In table \ref{tab2}, we note that the base model outperforms the small and medium variants. However, in Table \ref{tabindivtargets}, we observe that the Base model performs better on IPH, IVH, and SAH classes, whereas the Small model shows higher accuracy for epidural, SDH, and all classes. Based on these observations, we hypothesize that the improved performance observed on higher MLP dimensions indicates the ability of the model to encompass a larger amount of information and extract useful semantics for classification. However, the model shows signs of overfitting when the MLP dimension reaches 5120. Based on these results, we build our large \emph{Scopeformer (L)/8} model by adopting the configuration of the base variant with a global feature dimension $d = 1024$. The feature size increment resulted in a proportional increment of the model trainable parameters. The large model (L)/8 performed the best among the proposed variants. The improved performance observed on larger ViT sizes while increasing the input feature embedding space indicates richer information brought by these added features, where the model extracted useful semantics for classification. Increasing the feature space improved some of the classes at the expense of others, as evident from Table \ref{tabindivtargets}.

\subsection{The Effect of Number of ViT Encoders}
We evaluate the effect of the number of ViT encoders on the Scopeformer (L)/8 model using 4, 8, and 16 encoders. As presented in Table \ref{table-various-Scopeformer-designs} and Table \ref{tab2}, the number of parameters scales linearly with the number of encoders. We note that using 8 ViT encoders yields better results than a shallower model with 4 ViT encoders. However, a deeper model with 16 ViT encoders drastically reduces the model performance. We conclude that increasing the depth of the ViT model does not scale linearly and that there is a critical number of ViT encoders where the model performs optimally. 



\begin{figure}[H]
\begin{minipage}[b]{.45\linewidth}
  \centering
  \centerline{\includegraphics[width=4cm]{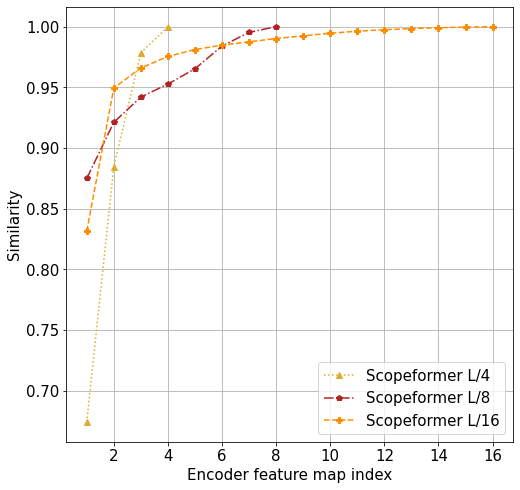}}
  \centerline{(a) Scopeformer variants}
\end{minipage}
\begin{minipage}[b]{.55 \linewidth}
  \centering
  \centerline{\includegraphics[width=4cm]{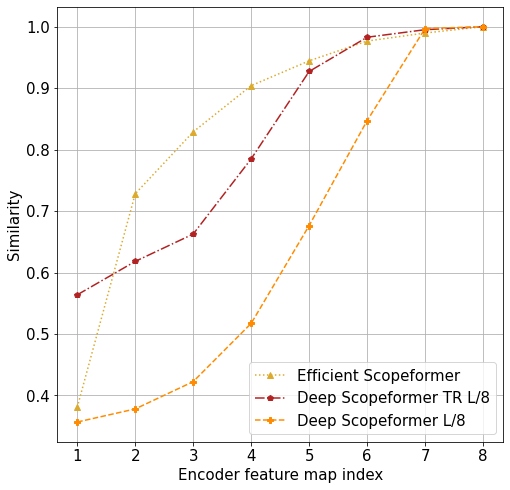}}
  \centerline{(b) Deep Scopeformer variants}
\end{minipage}
\caption{Cosine similarity of the ViT encoder feature maps with respect to the last encoder feature map. We observe the increased similarities across ViT encoder features function to the depth of Scopeformer models.}
\label{fig3}
\end{figure}
\unskip

In Figure \ref{fig3}, we plot the cosine similarity between the features generated by each ViT encoder and the last layer of the model. We observe that similarities across features generated by each ViT encoder rapidly increase for all proposed models. These similarities further increase in models with higher numbers of ViT encoders. We believe that the increased similarities among the features of the Scopeformer(L)/16 model may have contributed to the performance decline observed in Table \ref{tab2}. Similarly, reduced similarities among ViT features observed on Scopeformer(L)/4  may explain the observed sub-optimal performance. From these results, we conclude that the cosine similarity can be a good metric for model performance, as reduced or increased similarities may indicate sub-optimal performances of the Scopeformer model. Shallow models presenting reduced similarities may hint at higher performances by stacking more ViT layers, whereas deeper models may require additional data to reduce similarities across ViT features to perform optimally. The results also suggest that there is an optimum number of ViT encoders for the Scopeformer model based on the complexity of the dataset and the effectiveness of the convolution backbone networks.

\subsection{The Effect of Two Different Self-attention Variants}
The Deep Scopeformer (L)/8 builds on the Scopeformer (L)/8 model by replacing the MHSA layer with an MHRA layer. The additional trainable matrices $M$ add an insignificant number of parameters to the Scopeformer (L)/8 model. In Figure \ref{fig3} (b), we note substantial dissimilarities among ViT encoders' features for the \emph{Deep Scopeformer (L)/8} model. The result may imply an increased feature richness acquired by the model from the additional inter-correlations of the MHRA heads. This configuration resulted in an accuracy improvement by $+1.11\%$ as shown in Table \ref{tab4}.

\subsection{ViT Scopeformer Configurations}
We address the self-attention computational complexity problem by introducing a transpose layer before the ViT module. The attention weights matrix in \emph{Deep Scopeformer (L)/8} has a dimension of $1024^{2}$. In the second and third ViT configurations, the attention weights matrices have dimensions of $65^{2}$ and $64^{2}$, respectively. The use of the transpose layer has substantially contributed to the reduction of the number of trainable parameters as indicated in Table \ref{table-various-Scopeformer-designs}. This is due to the MHRA quadratic reduction in computation complexity. Additionally, transposing the input sequence effectively preserved the feature content retrieved by the feature extractor module and conserved the classification performance. Table \ref{tab4} shows the performance of the three proposed configurations. The proposed \emph{Efficient Scopeformer} variant performed relatively better than the \emph{Deep Scopeformer (L)/8} for a lower trainable parameter space. We speculate that the role of the ViT module in this configuration is to improve the global feature map that was previously optimized by the convolution backbone. The global feature map improvement resides in using attention computations to generate new features characterized by inter-correlations among all features generated by the convolution networks.

Our Efficient Transformer module improved the global features map correlations and contributed to better performance. We note that for the model \emph{Efficient Scopeformer (P)} pre-training the convolution block on the target dataset and freezing the entire block during training produces better performance than end-to-end training with around $30\%$  trainable parameters of the Efficient Scopeformer's convolution block. We argue that backbone CNNs and ViTs present different dynamics that require different model training settings.

\begin{table}[htbp]
\caption{Model performance on individual target classes}

\begin{tabularx}{0.45\textwidth}{X X X X X}
\toprule
\textbf{} & \multicolumn{4}{c}{\textbf{Accuracy}}            \\ \cline{2-5} 
 & \textbf{Large} & \textbf{Medium} & \textbf{Base} & \textbf{Small} \\ 
\midrule
\textbf{All}               & \textbf{71.34\%}  &  60.26\%           &70.5\% & 70.83\%          \\ 
\textbf{Epidural}          & 96.98\%         &  90.18\%          & 95.73\% & \textbf{98.08\%} \\ 
\textbf{IPH}  & 85.94\%          &  71.10\%  & \textbf{87.28\%} & 85.95\%          \\ 
\textbf{IVH}  & 90.5\%           & 70.73\%  & \textbf{91.72\%} & 90.13\%          \\ 
\textbf{SAH}    & \textbf{78.69\%}  & 65.49\%         & 78.57\%  & 77.04\%          \\ 
\textbf{SDH}     & \textbf{77.08\%}   & 60.78\%          & 74.35\% & 74.54\%        \\ 
\bottomrule
\end{tabularx}
\label{tabindivtargets}
\end{table}
\unskip

\begin{table}[htbp]
\caption{Model performance for different Scopeformer modalities}
\begin{tabularx}{0.5\textwidth}{X X X X}
\toprule
& \textbf{Accuracy} & \textbf{Loss}  &\textbf{Trainable Parameters}\\ 
\midrule
\textbf{Scopeformer (L)/8}  & 94.69\%           & 0.1197     &102M   \\ 
\textbf{Deep Scopeformer (L)/8}  & 96.03\%      & 0.1088     &102M   \\ 
\textbf{Deep Scopeformer TR (L)/8} & 95.40\%        & 0.1176 & \textbf{6M}        \\ 
\textbf{Efficient Scopeformer} & 95.77\%   & 0.1160 & \textbf{6M}   \\
\textbf{Efficient Scopeformer (P)} & \textbf{96.94}\%   & \textbf{0.0833} & \textbf{5M}   \\
\bottomrule
\end{tabularx}
\label{tab4}
\end{table}
\unskip

\subsubsection{Global and ViT Feature Maps}
\begin{figure}[htbp]
\begin{minipage}[b]{.48\linewidth}
  \centering
  \centerline{\includegraphics[width=4cm]{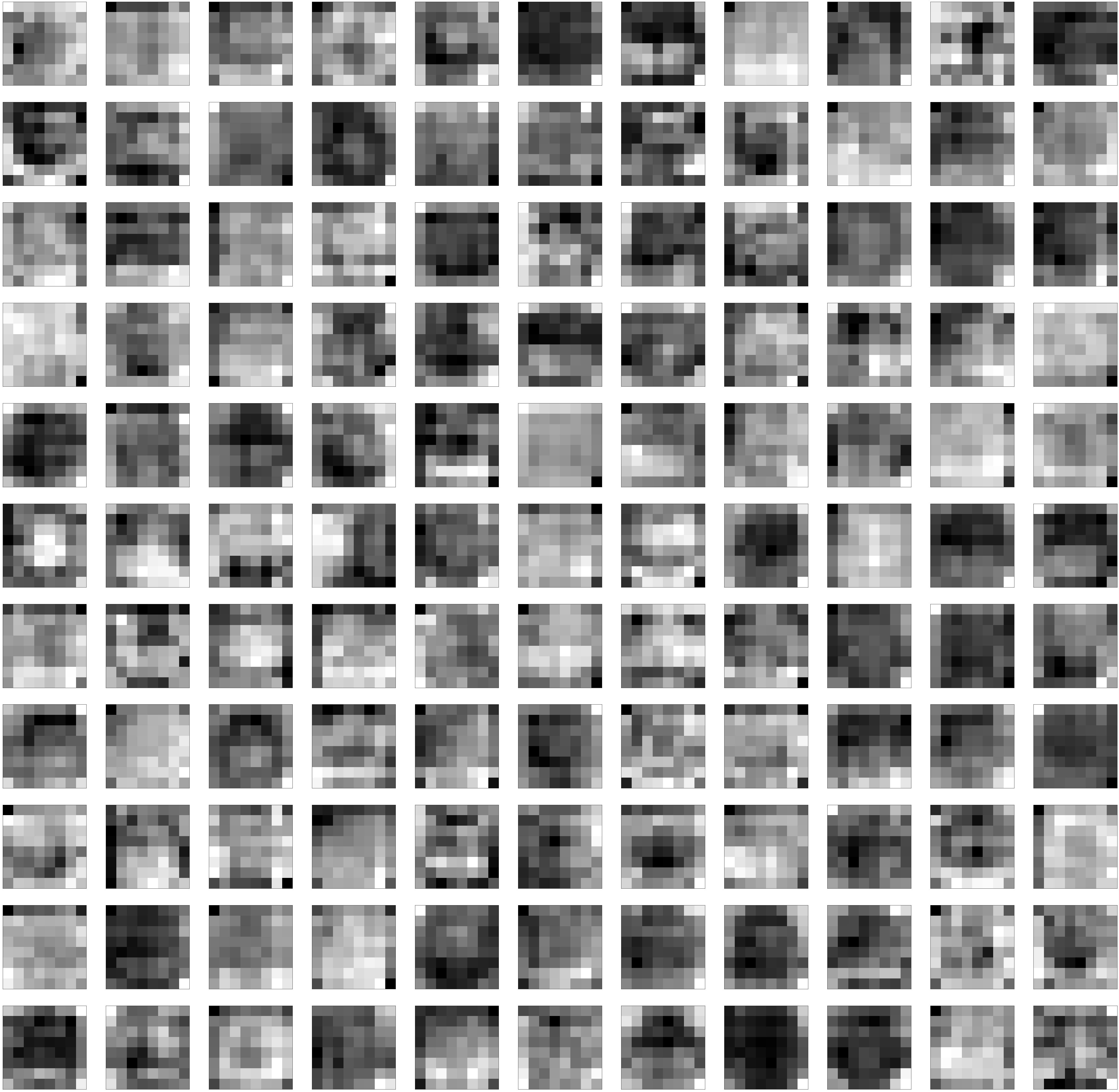}}
  \centerline{(a) Xception}\medskip 
\end{minipage}
\begin{minipage}[b]{.48\linewidth}
  \centering
  \centerline{\includegraphics[width=4cm]{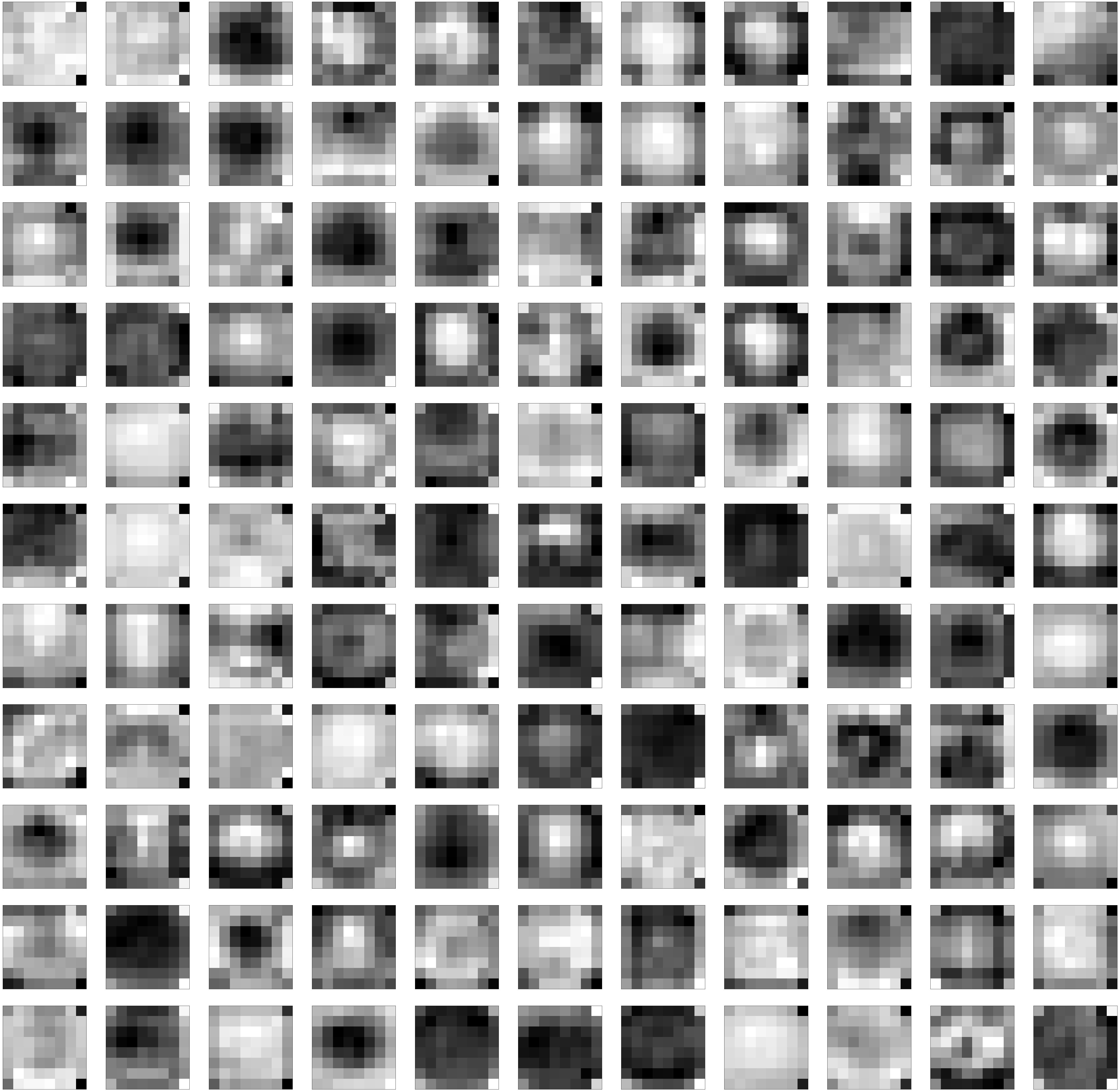}}
  \centerline{(b)EfficientNet B5}\medskip
\end{minipage}
\begin{minipage}[b]{.48\linewidth}
  \centering
  \centerline{\includegraphics[width=4cm]{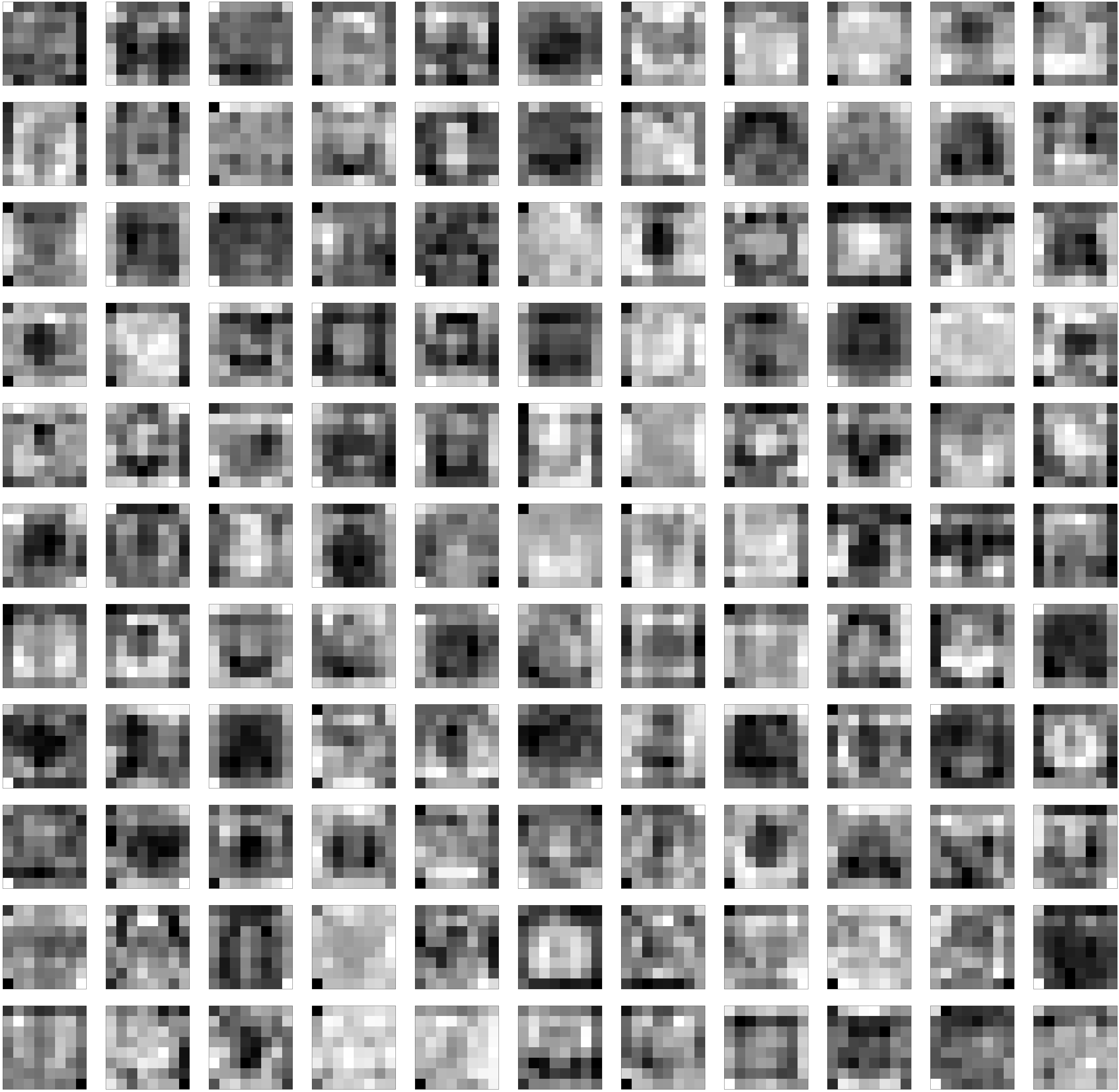}}
  \centerline{(c) DenseNet201 }\medskip
\end{minipage}
\hfill
\begin{minipage}[b]{0.48\linewidth}
  \centering
  \centerline{\includegraphics[width=4cm]{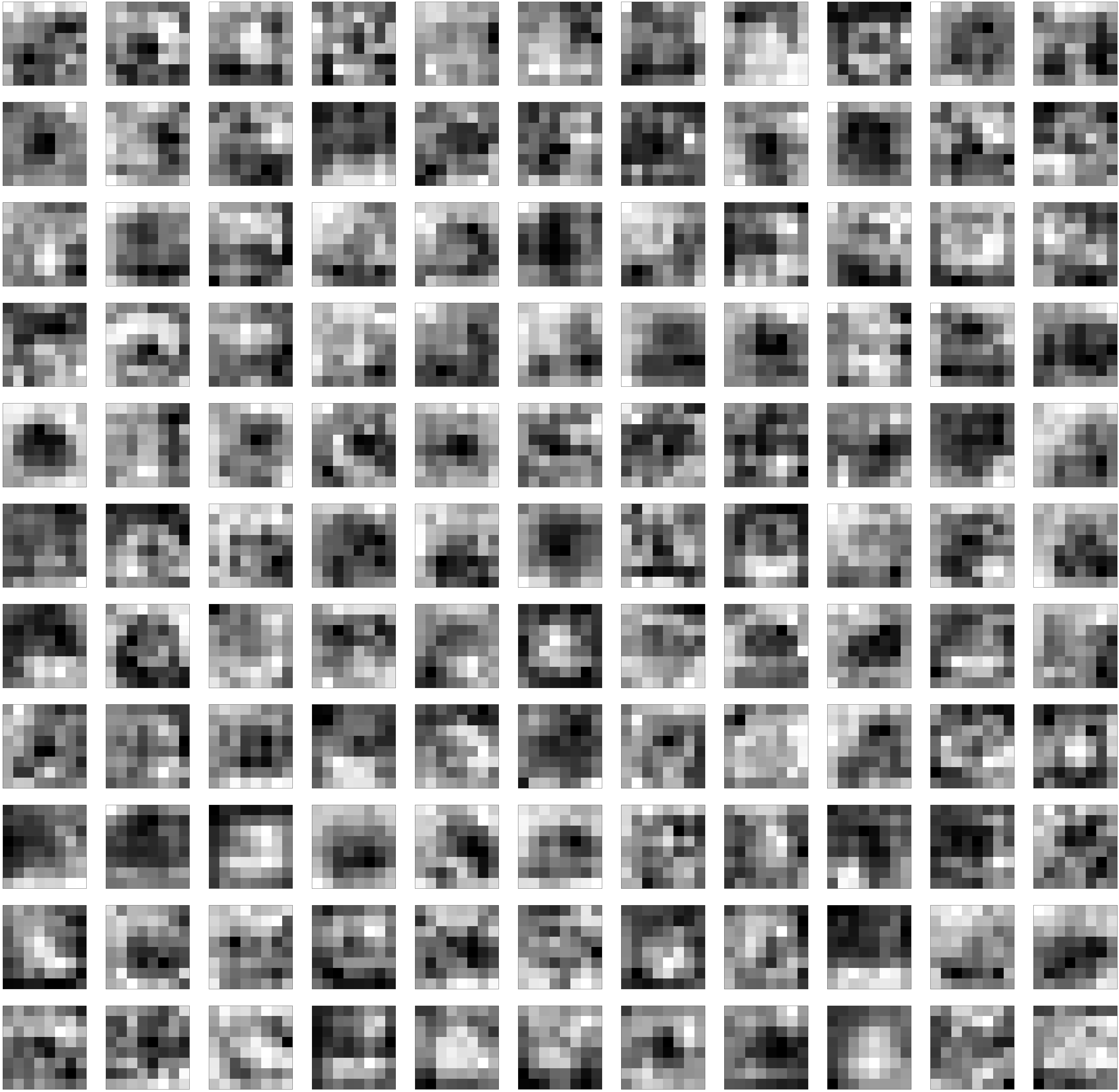}}
  \centerline{(d) ResNet152V2}\medskip
\end{minipage}
\caption{Feature maps visualization of an epidural type hemorrhage example. Scopeformer (L)/8}
\label{fig4}
\end{figure}

\begin{figure}[htbp]
\begin{minipage}[b]{.48\linewidth}
  \centering
  \centerline{\includegraphics[width=4cm]{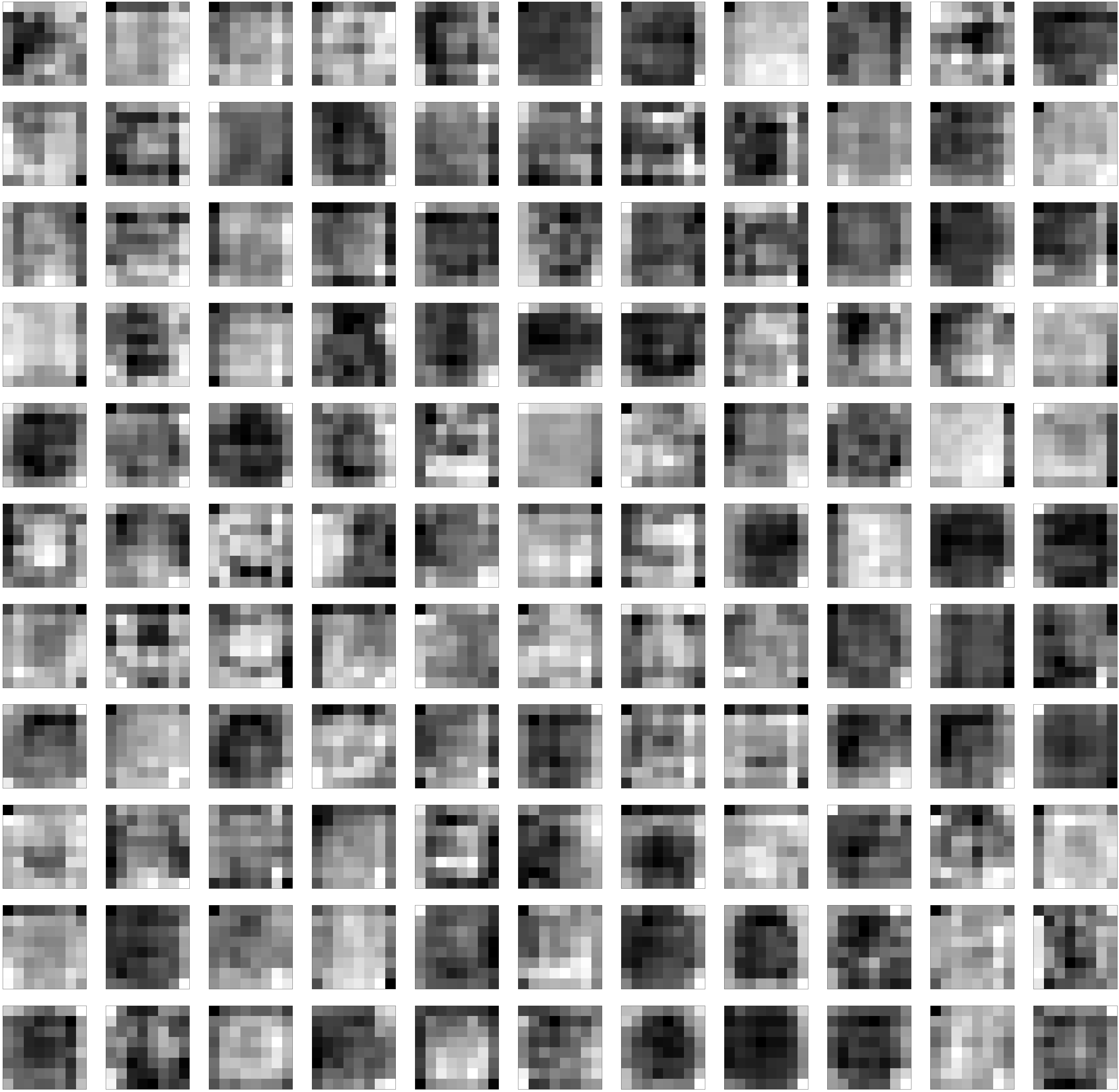}}
  \centerline{(a) Xception}\medskip
\end{minipage}
\begin{minipage}[b]{.48\linewidth}
  \centering
  \centerline{\includegraphics[width=4cm]{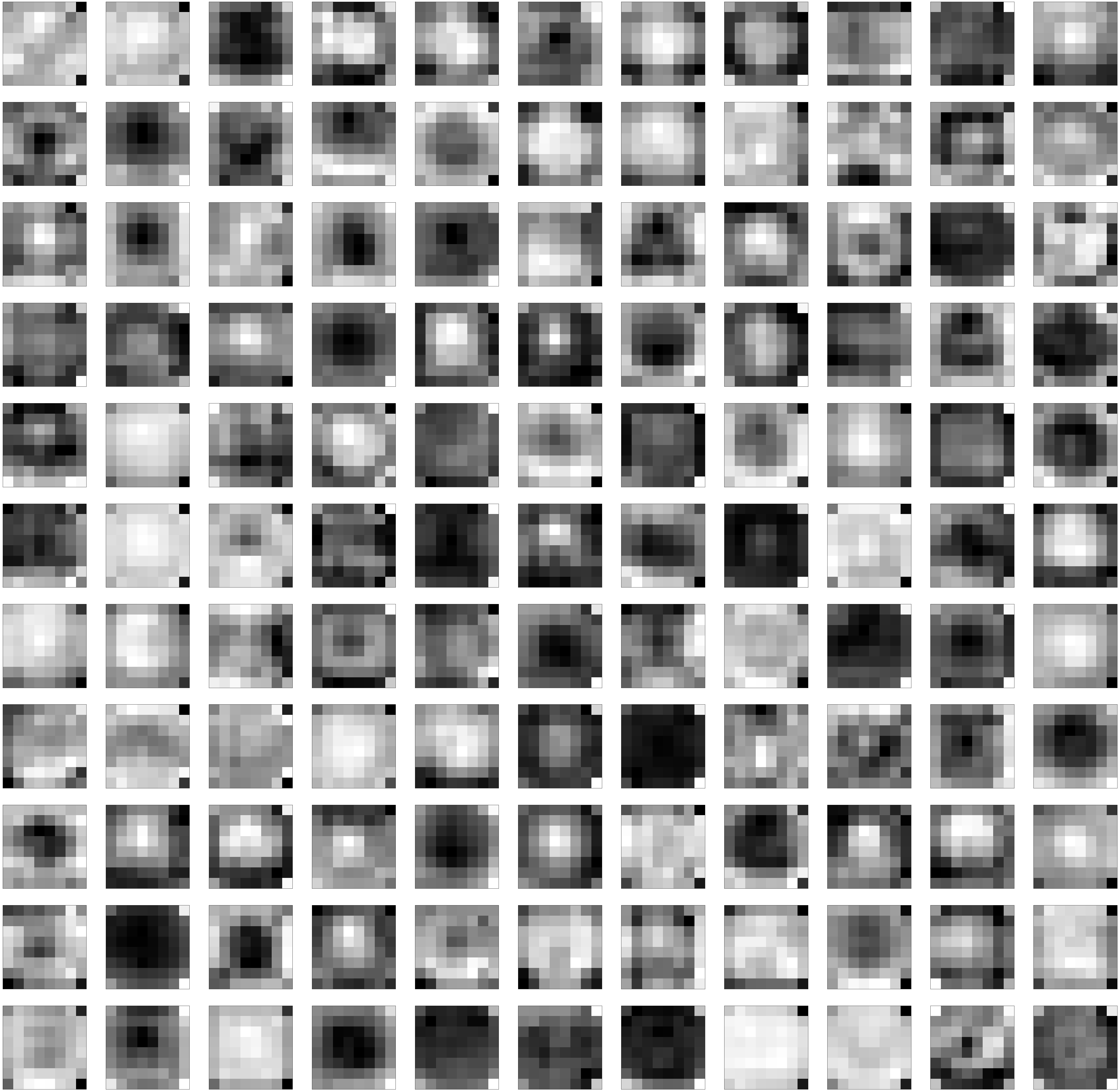}}
  \centerline{(b)EfficientNet B5}\medskip
\end{minipage}
\begin{minipage}[b]{.48\linewidth}
  \centering
  \centerline{\includegraphics[width=4cm]{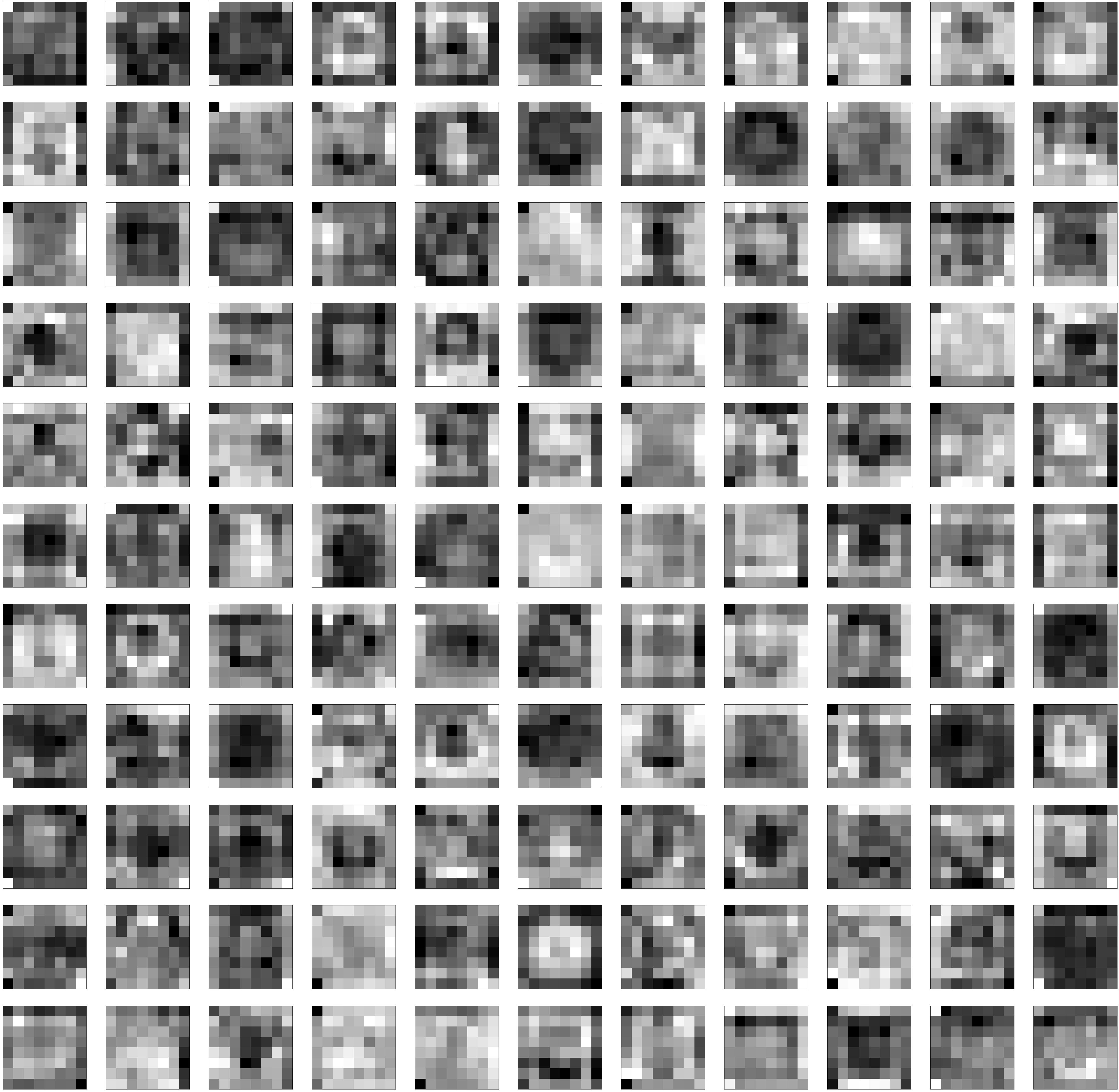}}
  \centerline{(c) DenseNet201 }\medskip
\end{minipage}
\hfill
\begin{minipage}[b]{0.48\linewidth}
  \centering
  \centerline{\includegraphics[width=4cm]{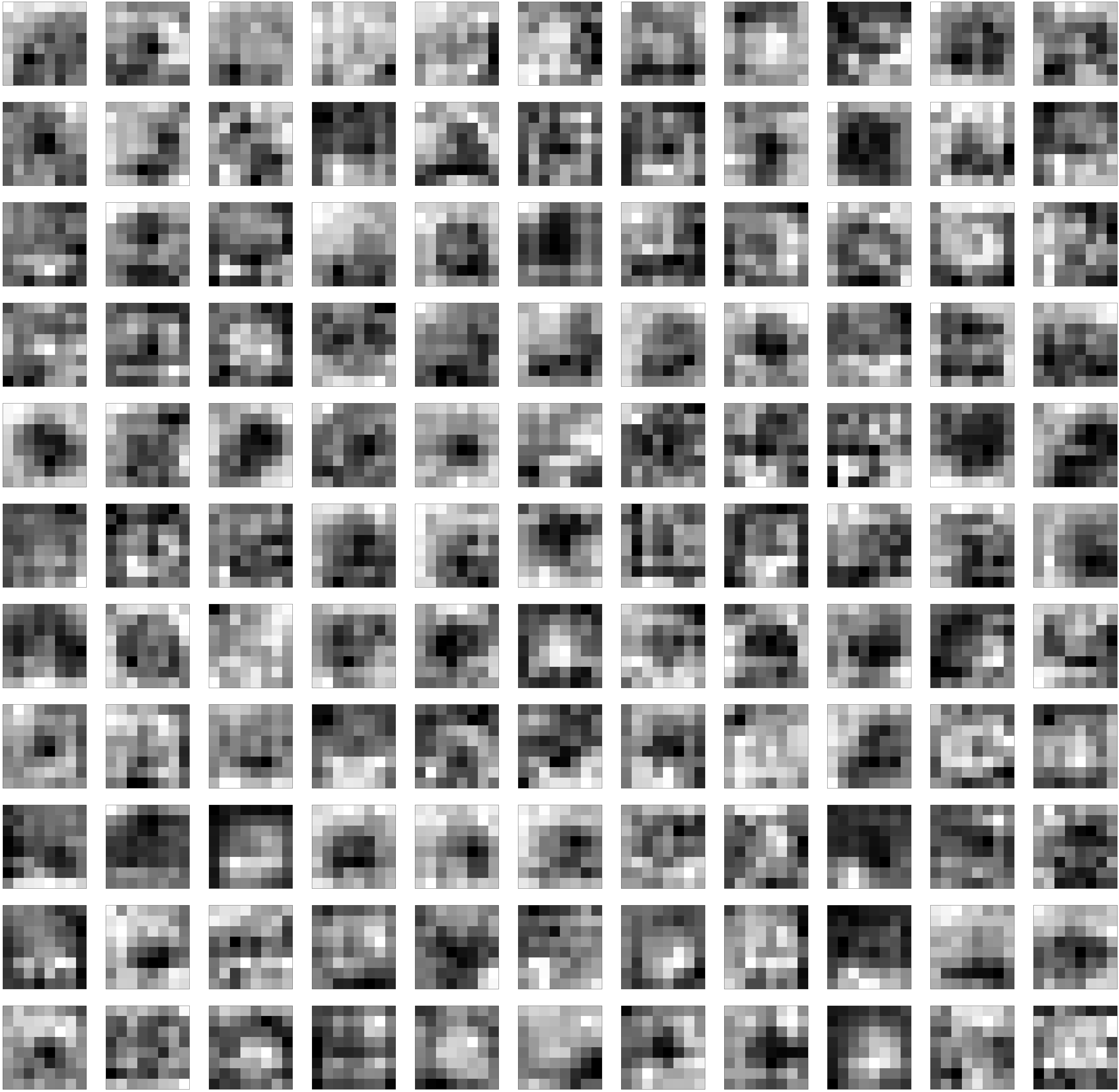}}
  \centerline{(d) ResNet152V2}\medskip
\end{minipage}
\caption{Feature maps visualization of an epidural type hemorrhage example. Deep Scopeformer (L)/8}
\label{fig5}
\end{figure}

\begin{figure}[htbp]
\begin{minipage}[b]{.48\linewidth}
  \centering
  \centerline{\includegraphics[width=4cm]{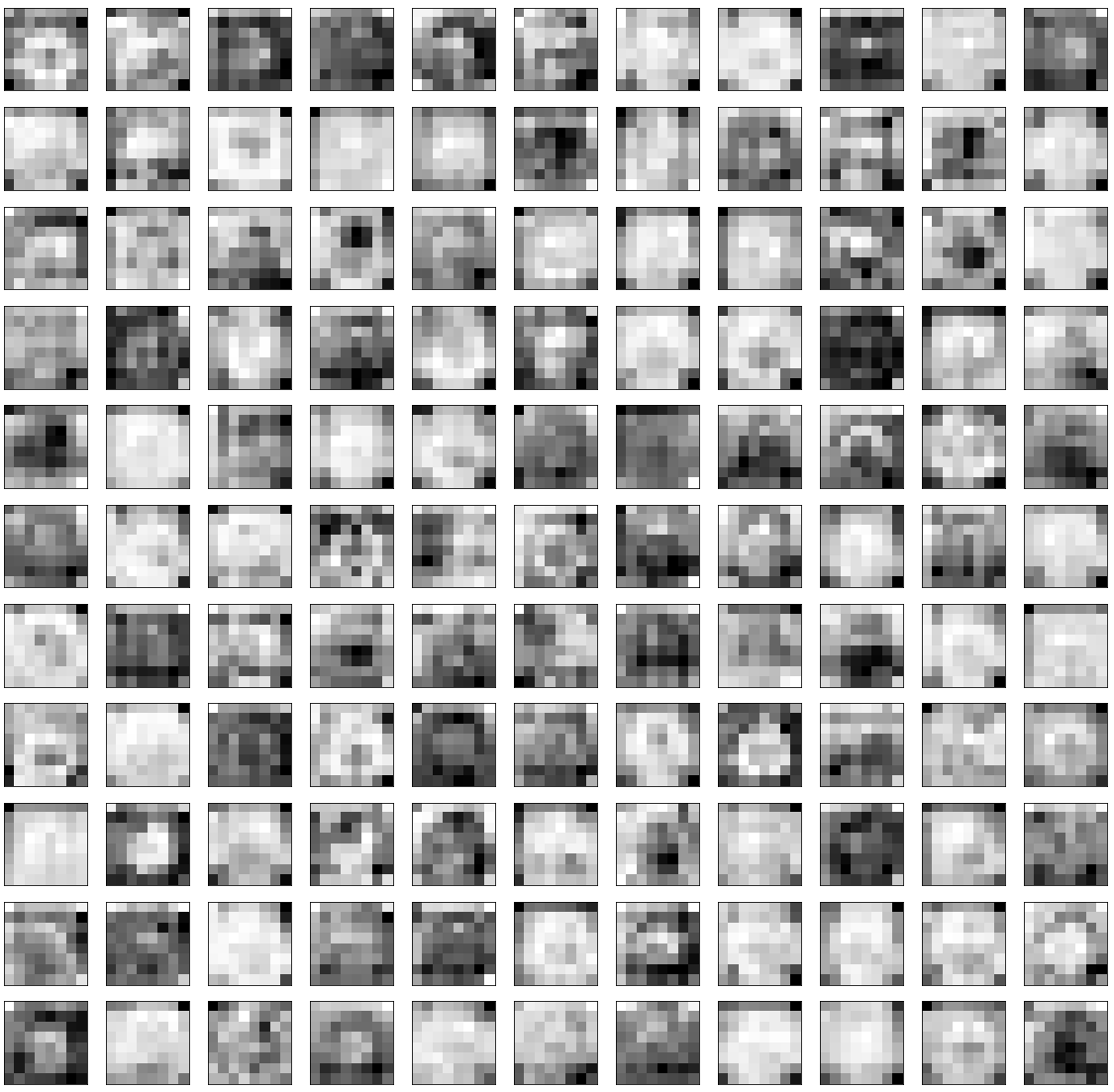}}
  \centerline{(a) Xception}\medskip
\end{minipage}
\begin{minipage}[b]{.48\linewidth}
  \centering
  \centerline{\includegraphics[width=4cm]{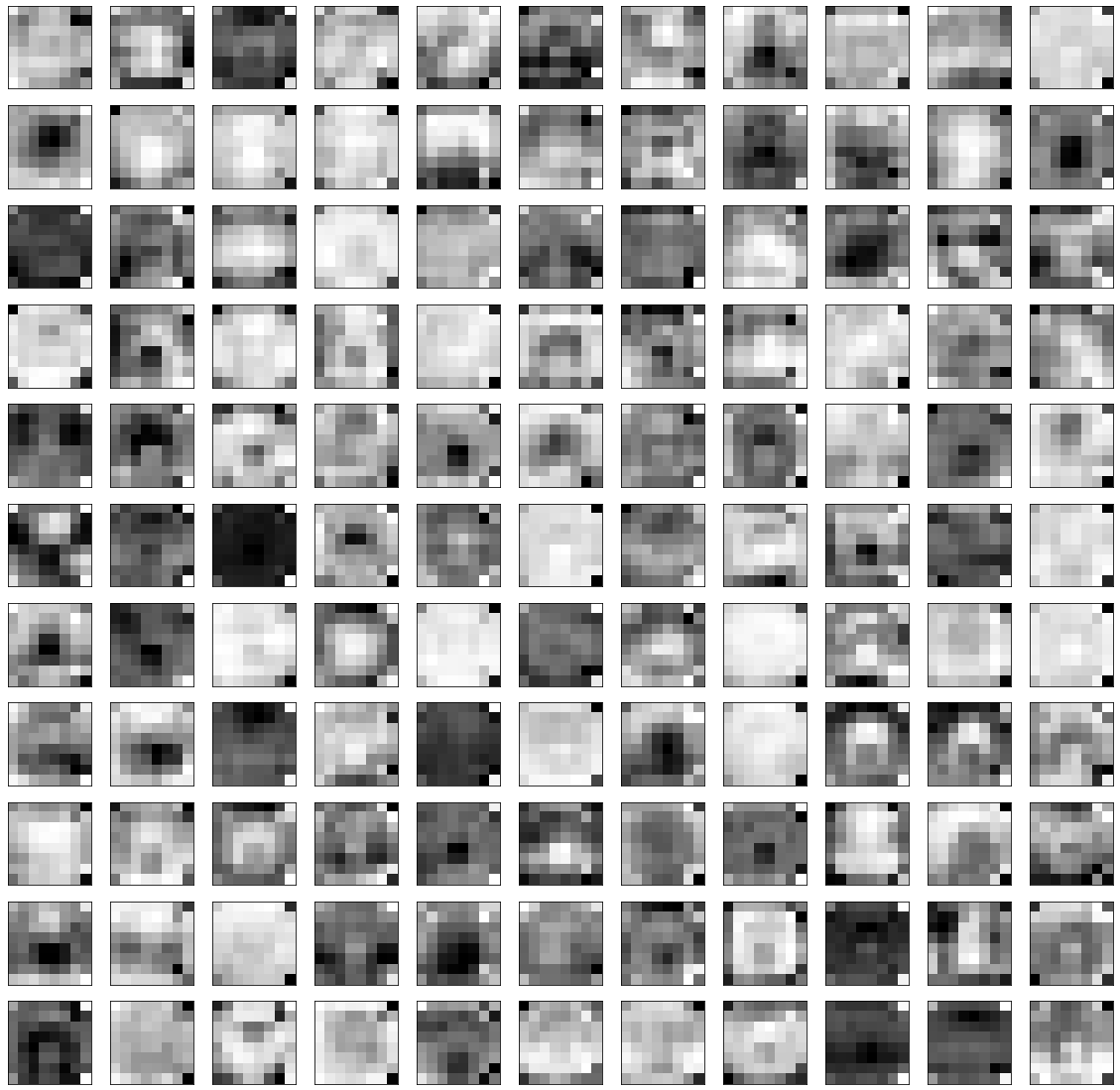}}
  \centerline{(b)EfficientNet B5}\medskip
\end{minipage}
\hfill
\begin{minipage}[b]{1.0\linewidth}
  \centering{\includegraphics[width=4cm]{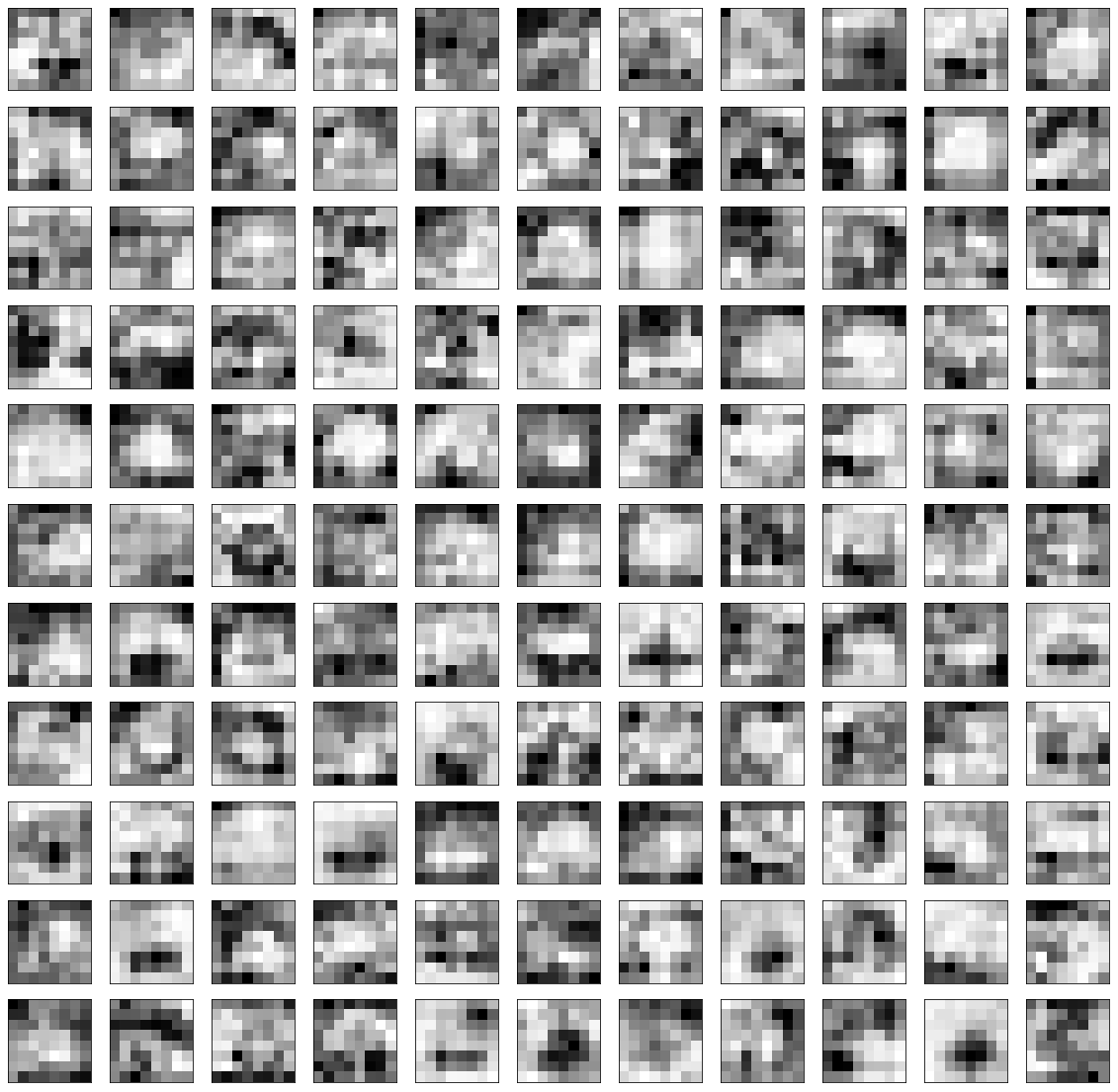}}
  \centerline{(c) ResNet152V2}\medskip
\end{minipage}
\caption{Feature maps visualization of an epidural type hemorrhage example. Efficient Scopeformer}
\label{fig6}
\end{figure}

Figures \ref{fig4}, \ref{fig5} and \ref{fig6} present convolution features generated by three Scopeformer architectures for an epidural example; Scopeformer (L)/8, Deep Scopeformer (L)/8, and Efficient Scopeformer. We observed high variability of the features generated by each CNN architecture. Furthermore, we observe that there is no apparent similarity among the features generated by different CNNs for all Scopeformer variants. Subsequently, the resultant global feature map has low redundancy and higher feature richness. However, among these models, we note that the DenseNet model showed the highest feature redundancy across the observed features. Therefore, we conducted an ablation study on the \emph{Deep Scopeformer TR}, which resulted in removing the DenseNet201 model from the \emph{Efficient Scopeformer} model backbone.

\subsubsection{Attention Patterns Visualizations}
\begin{figure}[htbp]

\begin{center}
\centering
    \includegraphics[width= 9 cm]{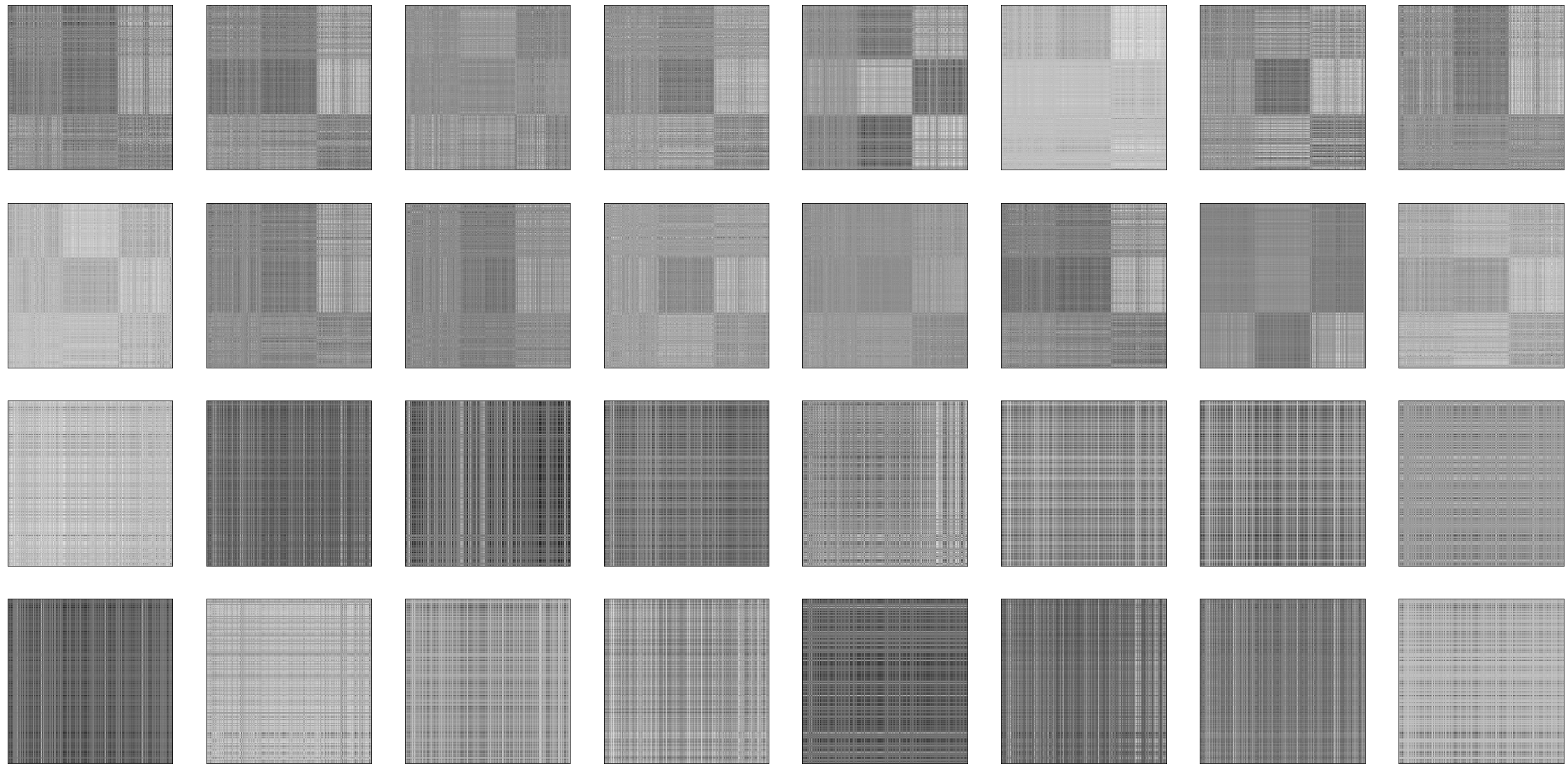}
     \caption{Attention pattern visualization of the Efficient Scopeformer model. The first and second rows represent the 16 attention heads of the first encoder layer. The third and fourth rows represent the 16 attention heads of the last encoder layer. Each attention map has a dimension of $384 \times 384$.}
      \label{attention_patterns}
\end{center}
\end{figure}

Figure \ref{attention_patterns} shows the attention patterns visualizations of the 16 MHRA heads concerning the first and last ViT encoders. In the first ViT layer, we observe that the model extracts high correlations among features derived from every CNN architecture. This observation suggests the high similarities among the input features of every CNN model. Each head learns different correlation patterns among the set of features.
However, deeper into the model, we observe that the model learns to extract global correlation patterns across all the CNN features. The generated set of features adds information about the relevance of every feature to the rest of the features, which contributes towards the observed higher performance.

\section{Conclusion}
We proposed a set of convolutional-based ViT models called Scopeformer to address the challenging problem of classifying types of hemorrhage in brain CT scans. We defined a range of model architectures for both CNNs and ViTs. We explored the effect of using multiple off-the-shelf CNN models on the global feature richness of the architecture and investigated a feature projection method to reduce the large redundant feature space into a lower and more efficient one. We conducted a parametric optimization study to evaluate the size effects on model performance and efficiency. We implemented three ViT configurations to evaluate the re-attention module within the Scopeformer model and the channel-wise versus feature-wise patch extraction of the global feature map. Results show increased richness of the features due to different CNN architectures. The re-attention module increased dissimilarities of ViT features resulting in improved performances and allowing deeper models. With our proposed feature-wise patch extraction method, the model size was reduced 17 times with comparable performance. Our Efficient Transformer module improved the global features map correlations and contributed to better performance. Furthermore, we observed that pre-training the convolution block on the target dataset and freezing the whole block during training produces better results than end-to-end training with $30\%$  trainable parameters of the Efficient Scopeformer's convolution block.

\section*{Acknowledgment}
This work was partly supported by the National Science Foundation Awards ECCS-1903466 and OAC-2008690.

\EOD
\end{document}